\documentclass[pdflatex,sn-nature]{sn-jnl}

\usepackage{graphicx}%
\usepackage{multirow}%
\usepackage{amsmath,amssymb,amsfonts}%
\usepackage{amsthm}%
\usepackage{mathrsfs}%
\usepackage[title]{appendix}%
\usepackage{xcolor}%
\usepackage{textcomp}%
\usepackage{manyfoot}%
\usepackage{booktabs}%
\usepackage{algorithm}%
\usepackage{algorithmicx}%
\usepackage{algpseudocode}%
\usepackage{listings}%
\usepackage{subcaption}%
\usepackage{hyperref}%
\usepackage{colortbl}
\usepackage[normalem]{ulem}

\newcommand{\methodname}{PROBE}
\newcommand{\methodfullname}{Post-hoc Reliability frOm Backbone Embeddings}
\newcommand{\papername}{Knowing when to trust machine-learned interatomic potentials}

\begin{document}

\title[{\methodname}]{\papername}

\author[1]{\fnm{Shams} \sur{Mehdi}}\email{shamsmeh@andrew.cmu.edu}

\author[1]{\fnm{Ilkwon} \sur{Cho}}\email{ilkwonc@andrew.cmu.edu}
\author*[1,2,3]{\fnm{Olexandr} \sur{Isayev}}\email{olexandr@olexandrisayev.com}

\affil[1]{\orgdiv{Department of Chemistry, Mellon College of Science}, \orgname{Carnegie Mellon University}, \orgaddress{\city{Pittsburgh}, \state{Pennsylvania, 15213}, \country{USA}}}

\affil[2]{\orgdiv{Department of Materials Science and Engineering}, \orgname{Carnegie Mellon University}, \orgaddress{\city{Pittsburgh}, \state{Pennsylvania, 15213}, \country{USA}}}

\affil[3]{\orgdiv{Ray and Stephanie Lane Computational Biology Department, School of Computer Science}, \orgname{Carnegie Mellon University}, \orgaddress{\city{Pittsburgh}, \state{Pennsylvania, 15213}, \country{USA}}}

\abstract{
Prevailing machine-learned interatomic potential (MLIP) uncertainty-quantification methods rely on ensembles of independently trained backbones. These methods scale unfavorably with foundation-scale MLIPs, and their member-disagreement signals correlate weakly with per-molecule prediction error. Here we probe the frozen per-atom representations of a pretrained MLIP with a compact discriminative classifier, recasting MLIP uncertainty quantification as \emph{selective classification} rather than error regression. The resulting method, {\methodname} (\methodfullname), produces a per-prediction reliability probability that monotonically tracks actual error without modification to the underlying model. Across large held-out evaluation sets and two structurally distinct MLIP architectures, {\methodname} outperforms ensemble disagreement as a binary reliability signal, which strengthens with the expressiveness of the backbone representation, implying a favorable scaling trajectory toward foundation-scale MLIPs. Multi-head self-attention additionally yields per-atom importance maps, providing chemically interpretable diagnostics at no additional computational cost. {\methodname} is post-hoc and architecture-agnostic, and is directly deployable on any MLIP that exposes per-atom representations.
}


\maketitle
\section{Introduction}\label{sec:intro}

Machine-learned interatomic potentials (MLIPs) are now widely used in computational chemistry and materials science~\cite{behler2007,kulik2022,schutt2018,batzner2022nequip,batatia2022mace,anstine2025aimnet2,deng2023chgnet,merchant2023gnome,wood2025uma}. They map atomic configurations to energies and forces using quantum mechanical reference data, achieving near density functional theory~\cite{hohenberg1964,kohn1965} (DFT) accuracy at a fraction of the cost. This enables large-scale atomistic simulations~\cite{mehdi2024enhanced} and accelerated materials discovery that would be intractable with first-principles methods. The field has advanced through  expressive model architectures, from local descriptors toward equivariant and attention-based message passing~\cite{batzner2022nequip,liao2024equiformerv2}, curation techniques spanning vast chemical spaces~\cite{anstine2025aimnet2,wood2025uma}, foundation model training~\cite{kovacs2023macemp,wood2025uma,qu2026allscaip}; broadening MLIP applicability across drug discovery~\cite{gomez2018automatic}, materials design, and automated reaction exploration.

Despite these successes, a persistent problem remains: how can a user know when to trust a given MLIP prediction? We address this question directly, by probing the internal representations of a trained MLIP with a lightweight classifier that learns to distinguish reliable from unreliable predictions. Uncertainties associated with MLIP predictions can be categorized into~\cite{hullermeier2021}: \emph{aleatoric} uncertainty, arising from irreducible noise in the data (e.g., DFT numerical errors or basis set incompleteness), and \emph{epistemic} uncertainty, arising from limited training data, distribution shift or unconverged training of the model. For MLIPs, the practically actionable source is epistemic: failures are systematic and typically tied to underrepresented chemical environments. Even within their training domain, MLIPs do not learn all chemical environments equally well; certain local bonding patterns, rare element combinations, or strained geometries are systematically harder to fit. Long-range intermolecular interactions are fundamentally harder to extrapolate than intramolecular ones. Examples include $\pi-\pi$ stacking, charge transfer, and halogen bonding driven by anisotropic $\sigma$-holes (e.g., in Br and I). These physics-driven limitations produce large prediction errors that can corrupt high-throughput screening pipelines, active learning loops, and geometry optimization workflows before being noticed~\cite{schran2020,smith2018,tan2023}. Perez et al.~\cite{perez2025misspecified} recently showed this concretely for the MACE-MPA-0 foundation MLIP on tungsten defects: ensemble disagreement captures epistemic variance but misses model misspecification. Against this backdrop, uncertainty quantification (UQ) has become an explicit research priority for the atomistic-ML community~\cite{grasselli2025uqera, heid2023}.

The dominant approach to MLIP UQ has been the committee or deep ensemble~\cite{lakshminarayanan2017,smith2018}: several independently trained models whose output variance serves as a proxy for uncertainty~\cite{schran2020,smith2018}. Committee potentials have driven much of the progress in uncertainty-guided active learning~\cite{smith2018,vandermause2020,podryabinkin2017,kulichenko2023}, including uncertainty-driven molecular dynamics~\cite{zaverkin2024ubmd} and hyperactive learning strategies~\cite{vdoord2023hal} that bias sampling toward high-uncertainty regions. However, Kurniawan et al.~\cite{kurniawan2025comparative} recently showed that ensemble uncertainty in out-of-distribution regimes can plateau or even decrease as predictive errors grow. For equivariant neural network potentials specifically, Lu et al.~\cite{lu2023equivariant} found that ensemble uncertainty estimates are overconfident and poorly predictive of actual errors, suggesting the problem is not merely a matter of method but of a fundamental mismatch between ensemble disagreement and true error. Additionally, there are practical drawbacks of running $N$ models which costs $N$ times as much in both compute and memory. Recent UQ work on foundation-scale MLIPs that scale to tens of millions of parameters, therefore abandons full ensembling in favorof readout-layer or shared-backbone alternatives~\cite{bilbrey2025foundationuq}.

The community has pursued a range of approaches involving single model MLIPs to avoid the ensemble overhead. Monte Carlo Dropout~\cite{gal2016} applied to MLIPs~\cite{wen2020} provides cheap stochastic estimates, though calibration often degrades under distribution shift. Gaussian process~\cite{williams2006gaussian} force fields~\cite{bartok2022gpr} such as FLARE~\cite{vandermause2020,vandermause2020gpr} offer principled Bayesian posteriors but scale poorly to large training sets. Bayesian neural network potentials~\cite{farris2025} and Bayesian learned interatomic potentials (BLIPs)~\cite{coscia2025blips} can yield well-calibrated uncertainties but require modifying or re-training the backbone with variational inference. Evidential deep learning for interatomic potentials~\cite{xu2024evidential} and loss-trajectory analysis~\cite{vita2024ltau} represent newer single-network approaches that avoid additional training but require non-standard training procedures.  Conformal prediction has recently been adapted for MLIPs~\cite{ho2025}, giving distribution-free coverage guarantees without backbone modification, which produces prediction intervals rather than reliability signals. Multi-head committee models for MACE~\cite{beck2025} share a backbone to reduce cost, but still require training and maintaining additional output heads.

A closely related line of work avoids ensembling by extracting uncertainty signals from a single backbone's internal representations. Janet et al.~\cite{janet2019quantitative} and, more recently, Musielewicz et al.~\cite{musielewicz2024latent} used latent-space distance to the training set as a quantitative error metric, the latter on a frozen equivariant GNN potential. Bigi et al.~\cite{bigi2024prediction} formalised this intuition as a post-hoc \emph{prediction rigidity} scalar. It measures how sensitive a trained network's output is to its training data, without additional sampling. Kellner and Ceriotti~\cite{kellner2024shallow} augmented a shared backbone with lightweight \emph{shallow-ensemble} heads, recovering much of a full ensemble's signal at a small fraction of the cost. A previous comparative study~\cite{tan2023} concluded that, across this diverse landscape, no single-model UQ method consistently outperforms ensembles for MLIP error estimation.

Here, we argue that part of the difficulty with MLIP UQ comes from an overly ambitious problem formulation. Most existing methods try to predict the \emph{magnitude} of the error, a nonlinear regression problem over a heavy-tailed distribution. We instead ask a simpler question: is this prediction reliable? This binary framing captures most of the practical information a UQ system needs to provide. Downstream decisions are themselves binary: accept a geometry optimization, include a molecule in a training set, or trigger a DFT recalculation. 

\begin{figure}[h]
\centering
\includegraphics[width=\textwidth]{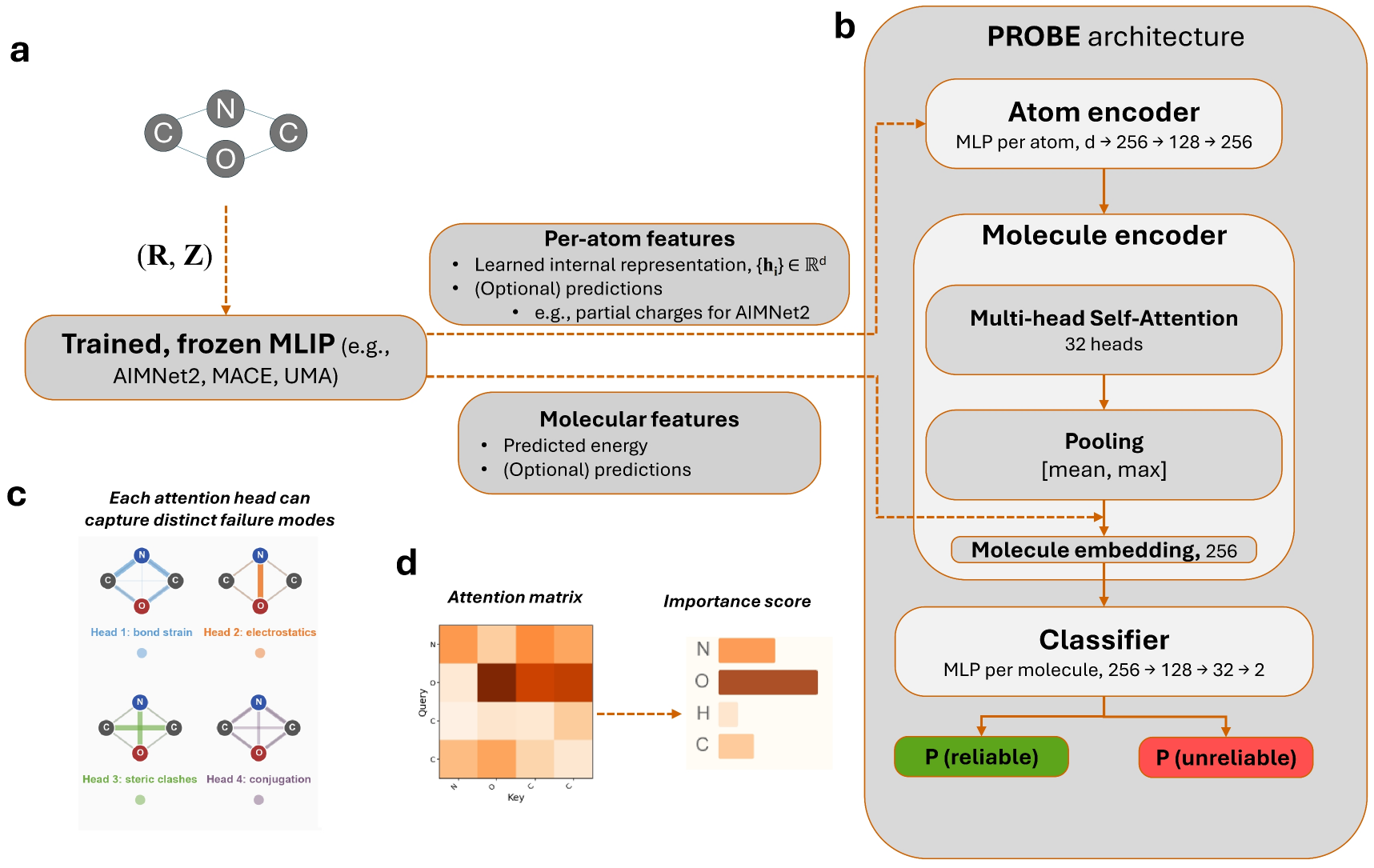}
\caption{\textbf{{\methodname} architecture overview.}
\textbf{(a)} A frozen, pre-trained MLIP processes
atomic coordinates $(\mathbf{R}, \mathbf{Z})$ and exposes per-atom latent
representations $\{h_i\} \in \mathbb{R}^d$ alongside molecular-level predictions.
{\methodname} attaches to these frozen representations without modifying the
underlying MLIP.
\textbf{(b)} {\methodname} architecture. An atom encoder projects per-atom features into a fixed-dimensional space. A molecule encoder
applies multi-head self-attention followed by mean and max pooling to produce a
fixed-size molecular embedding. A classifier MLP outputs $P(\text{reliable})$ and $P(\text{unreliable})$.
\textbf{(c)} Schematic illustration of the multi-head attention mechanism:
Different heads can, in principle, attend to distinct structural contexts within
a molecule (e.g., strained bonds, polar regions, steric clashes, or conjugated
systems), producing complementary contributions to the molecular embedding.
Specific head specialization shown is not a claim of the present work.
\textbf{(d)} Per-atom importance scores derived from the attention matrix by summing
attention weights received by each atom (Eq.~\ref{eq:importance}), enabling
identification of the specific atoms driving an unreliable prediction.}\label{fig:architecture}
\end{figure}

We present {\methodname} ({\methodfullname}), a lightweight framework built on this idea. {\methodname} attaches a small classifier to the frozen internal representations of a trained MLIP. The classifier is trained to distinguish reliable from unreliable predictions using per-atom embeddings from the backbone, without modifying the backbone itself. We hypothesize that a trained MLIP's internal representations already encode whether a chemical environment was well-covered during training, and that a simple classifier can exploit this signal at inference time.

This reframing connects MLIP UQ to the literature on \emph{selective prediction} and \emph{confidence-based abstention}\cite{elyaniv2010selective, geifman2017selective, geifman2019selectivenet, devries2018confidence}, in which a model identifies inputs whose predictions should be trusted versus deferred. Representation-based reliability scoring is established for out-of-distribution detection in computer vision, language modeling\cite{hendrycks2017baseline, lee2018mahalanobis, sun2022knnood, yang2024oodsurvey}, and scientific ML~\cite{fannjiang2022conformal}, but has not been applied to MLIP energetics, where ensemble disagreement remains the default. Methodologically, {\methodname} is a \emph{probing classifier} in the sense of Alain and Bengio~\cite{alain2016probing}: a small auxiliary model trained on a frozen backbone's intermediate features, exploiting the observation that a trained backbone's internal representations already encode whether a chemical environment was well-covered during training. Figure~\ref{fig:architecture} summarizes {\methodname} workflow, which has the following key features:

\begin{enumerate}
\item \textbf{Internal MLIP representations for atomic embedding}: {\methodname} operates directly on the per-atom embeddings produced by a single, trained MLIP backbone, capturing rich chemical environment information without any modification to the underlying model.

\item \textbf{Multi-head self-attention for molecular embedding}: Atom-level embeddings are processed through a self-attention module to construct a molecular embedding, enabling {\methodname} to capture both local and global chemical context while naturally handling variable-sized molecules. Here, the attention mechanism handles the non-additive nature of prediction error, with different heads learning to specialize on distinct failure modes present in the training data.

\item \textbf{Data-driven class boundaries}: The reliability threshold is defined as a percentile of the training error distribution; we use the 50th percentile (balanced classes) throughout this work. In Section~\ref{sec:discussion} we discuss why moving off the median degrades performance, and why raising the probability cutoff on a balanced model is the better way to target severe outliers.

\item \textbf{Interpretable atom importance}: Attention weights provide chemically meaningful, per-atom reliability maps without any additional computation, identifying which atoms are responsible for anomalously large prediction errors.

\item \textbf{Architecture agnosticism}: {\methodname} requires only access to per-atom internal representations and can therefore be deployed as a post-hoc module on most modern MLIP architectures, demonstrated here for both AIMNet2~\cite{anstine2025aimnet2} and MACE~\cite{batatia2022mace}.
\end{enumerate}

We tested {\methodname} on AIMNet2 and MACE because they use contrasting representation families: chemically informed atom-in-molecule vectors for AIMNet2~\cite{anstine2025aimnet2}, and equivariant graph-based node embeddings for MACE~\cite{batatia2022mace, kovacs2023macemp}, which lets us evaluate whether {\methodname} generalizes across architectures. Details about {\methodname} architecture, training procedure and complete algorithm is provided in Methods (Section \ref{sec:methods}).

\section{Results}\label{sec:results}
\subsection{{\methodname} achieves high-precision reliability classification for AIMNet2}\label{sec:aimnet2}

We trained {\methodname} on top of AIMNet2~\cite{anstine2025aimnet2} that produces 256-dimensional per-atom AIM embeddings that encode the local chemical environment of each atom through three rounds of message passing. The model outputs per-atom partial charges, total energy both of which are passed to {\methodname} in addition to the AIM vectors. Results are reported on an additional held-out test set of 3.76M molecules absent from both AIMNet2 and {\methodname} training set. Additional details on AIMNet2 training and evaluation are provided in Methods (Section \ref{sec:aimnet2_setup}).

We evaluate {\methodname} along three standard UQ axes. \emph{Discrimination} measures how well the reliability score separates truly reliable from unreliable predictions; we report classification accuracy, Matthews correlation coefficient (MCC), and F1. \emph{Ranking} measures how well the score orders molecules by actual error magnitude, captured below by the monotonic relationship between $P(\text{unreliable})$ and mean error. A third axis, \emph{calibration} (agreement between predicted probabilities and empirical frequencies in each probability bin), is a distinct property that we also characterize.

\begin{figure}[h]
\centering
\includegraphics[width=\textwidth]{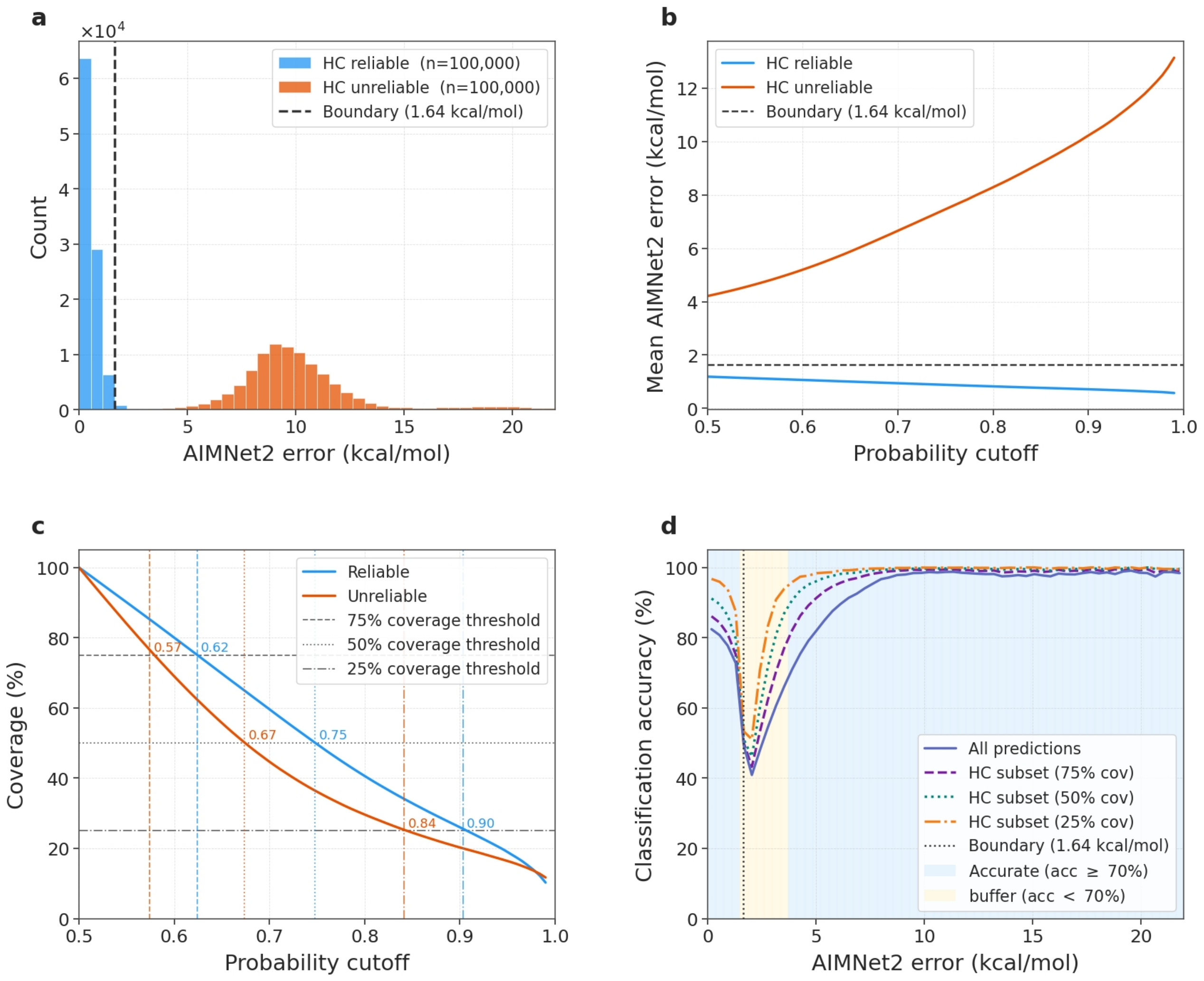}
\caption{\textbf{{\methodname} performance on AIMNet2 (3.76M held-out molecules, 50th-percentile boundary at 1.64~kcal/mol).}
(a) Error distributions for the top-100,000 high-confidence reliable (blue) and high-confidence unreliable (orange) predictions, with the class boundary shown as a dashed line.
(b) Mean AIMNet2 error as a function of probability cutoff for the reliable (blue) and unreliable (orange) classes.
(c) High-confidence coverage as a function of probability cutoff for both classes. Vertical and horizontal markers indicate cutoffs corresponding to 25\%, 50\%, and 75\% coverage thresholds.
(d) Overall classification accuracy (solid navy) and HC subset accuracy (dashed lines, one per coverage threshold) as a function of binned AIMNet2 error magnitude. Shading indicates accurate (blue, overall accuracy $\geq 70\%$) and buffer (yellow, overall accuracy $< 70\%$) regions.}\label{fig:aimnet2_results}
\end{figure}

\textbf{Error separation at high confidence.} Fig.\ref{fig:aimnet2_results}(a) shows error distributions for the top-100,000 HC-reliable and HC-unreliable predictions ($\sim$2.7\% of the test set on each side). High-confidence (HC)-reliable predictions have mean error 0.494kcal/mol while HC-unreliable reach 11.319kcal/mol, well separated on either side of the 1.64kcal/mol boundary, confirming that {\methodname}'s confidence signal isolates genuinely distinct error regimes rather than partitioning noise near the boundary. The two HC subsets occupy disjoint regions of the error distribution on either side of the class boundary, indicating that {\methodname}'s confidence signal isolates genuinely distinct error regimes rather than merely partitioning noise near the boundary. Table~\ref{tab:aimnet2_hc} reports analogous means at fixed probability cutoffs covering larger fractions of the test set. 

\begin{table}[h]
\caption{\textbf{{\methodname} high-confidence performance for AIMNet2 (50th-percentile boundary, 1.64~kcal/mol).} Evaluated on 3.76M held-out molecules. Coverage = fraction of test set assigned a high-confidence label at the given cutoff. $\bar{e}_\text{rel}$ and $\bar{e}_\text{unrel}$ are mean AIMNet2 errors (kcal/mol) for the HC-reliable and HC-unreliable subsets respectively. Matthews correlation coefficient (MCC), and F1 score are also reported.}\label{tab:aimnet2_hc}
\small
\begin{tabular*}{\columnwidth}{@{\extracolsep\fill}lcccccc}
\toprule
Cutoff & Coverage & Acc. & MCC & F1 & $\bar{e}_\text{rel}$ & $\bar{e}_\text{unrel}$ \\
 & & & & & (kcal/mol) & (kcal/mol) \\
\midrule
All ($P \geq 0.5$) & 100.0\% & 71.6\% & 0.402 & 0.633 & 1.19 & 4.21 \\
$P \geq 0.6$       &  75.9\% & 77.2\% & 0.509 & 0.685 & 1.06 & 5.20 \\
$P \geq 0.7$       &  54.0\% & 83.2\% & 0.628 & 0.749 & 0.94 & 6.66 \\
$P \geq 0.8$       &  36.5\% & 88.7\% & 0.746 & 0.826 & 0.82 & 8.29 \\
$P \geq 0.9$       &  23.9\% & 93.2\% & 0.849 & 0.898 & 0.71 & 10.22 \\
$P \geq 0.95$      &  18.0\% & 95.5\% & 0.902 & 0.936 & 0.65 & 11.50 \\
$P \geq 0.99$      &  10.8\% & 97.9\% & 0.957 & 0.974 & 0.58 & 13.14 \\
\bottomrule
\end{tabular*}
\end{table}

\textbf{Confidence monotonically tracks error.} Fig.\ref{fig:aimnet2_results}(b) shows that mean error for reliable predictions decreases monotonically and for unreliable predictions increases monotonically as the probability cutoff increases for creating a subset (molecules with predicted class probability lower than this cutoff are ignored), across the full range $P \in [0.5, 1.0]$. At $P \geq 0.99$, HC-reliable mean error reaches 0.58kcal/mol while HC-unreliable reaches 13.14kcal/mol (Table\ref{tab:aimnet2_hc}). Calibration analysis (SI Fig. S7a) confirms excellent ECE = 0.039 and Brier score = 0.182 for the test dataset, indicating that $P(\text{unreliable})$ reflects empirical frequencies rather than a mere rank-ordering signal.

\textbf{Coverage-accuracy tradeoff.} Fig.~\ref{fig:aimnet2_results}(c) shows the selective-prediction operating characteristic: monotonic coverage and accuracy trade off with the probability cutoff. Table~\ref{tab:aimnet2_hc} reports combined accuracy, MCC, F1, and mean errors across a range of operating points. At $P \geq 0.7$, {\methodname} retains 54.0\% of the test set at 83.2\% accuracy; at $P \geq 0.9$, coverage drops to 23.9\% while accuracy rises to 93.2\%; at $P \geq 0.99$, 10.8\% of molecules are retained at 97.9\% accuracy. All relevant metrics improve monotonically with probability cutoff, indicating that high-confidence predictions are more likely to be correct. Even at strict cutoffs, {\methodname} still yields high-confidence predictions for a sizeable fraction of the test set.

\textbf{Accuracy as a function of error magnitude.} Fig.\ref{fig:aimnet2_results}(d) shows classification accuracy binned by AIMNet2 error. Near the 1.64kcal/mol boundary (buffer region, shaded yellow) the problem is inherently ambiguous and accuracy drops below 70\% overall; away from it, classification is confidently correct on both sides (shaded blue). HC subset accuracy curves lie above the overall curve throughout the buffer and the buffer narrows at stricter cutoffs, confirming that confidence filtering preferentially removes ambiguous near-boundary molecules.

Overall, {\methodname} achieves 71.6\% accuracy, F1 = 0.633, and MCC = 0.402 on the full 3.76M test set (Table~\ref{tab:aimnet2_hc}). Confidence filtering converts this into a highly precise classifier: at $P \geq 0.9$, accuracy = 93.2\%, F1 = 0.898, and MCC = 0.849.

\subsection{Comparison with ensemble-based UQ}\label{sec:ensemble}

The standard UQ approach for MLIPs uses the standard deviation of an ensemble of independently trained models as a proxy for prediction uncertainty~\cite{lakshminarayanan2017,schran2020,smith2018}. We benchmark this approach using a 4-model AIMNet2 ensemble trained on the same data and evaluated on the same 3.76M molecule test set, scaling ensemble standard deviation by $1/\sqrt{N_\mathrm{atoms}}$ to reduce the effect of molecule size. The Spearman correlation between scaled $\sigma$ and the actual AIMNet2 error is $\rho = 0.229$, a weak monotonic signal at the individual-molecule level, consistent with prior observations~\cite{lu2023equivariant} that ensemble members trained on the same data make correlated mistakes in chemically underrepresented regions. To benchmark ensemble $\sigma$ as a reliability \emph{classifier} (the quantity {\methodname} is optimized for), we threshold scaled $\sigma$ at its median (SI Fig.~S2) to define reliable and unreliable classes. This protocol evaluates $\sigma$ as a classifier under the same decision framing as {\methodname}.

As a baseline, we use majority-class prediction: the class boundary of 1.64~kcal/mol produces a 60/40 reliable/unreliable split on the test set, so always predicting the reliable class yields 60\% accuracy.

\begin{table}[h]
\caption{\textbf{Comparison of UQ methods for AIMNet2 (boundary at 1.64~kcal/mol).} Evaluated on 3.76M held-out molecules. The majority-class baseline always predicts reliable, reflecting the 60/40 class split on the test set. HC = high-confidence subset; not applicable (N/A) for methods that do not produce calibrated probabilities.}\label{tab:ensemble_compare}
\small
\begin{tabular*}{\columnwidth}{@{\extracolsep\fill}lccc}
\toprule
Method & Accuracy & HC Accuracy & HC Accuracy \\
 & (all) & ($P \geq 0.7$) & ($P \geq 0.9$) \\
\midrule
Majority-class baseline & 60.0\%          & N/A             & N/A \\
Ensemble std (4 models) & 57.6\%          & N/A             & N/A \\
\rowcolor{blue!10}{\methodname} (ours) & \textbf{71.6\%} & \textbf{83.2\%} & \textbf{93.2\%} \\
\bottomrule
\end{tabular*}
\end{table}

Table~\ref{tab:ensemble_compare} compares the majority-class baseline, ensemble thresholding, and {\methodname}. When ensemble standard deviation is thresholded at its median to produce a binary classifier, it achieves 57.6\% accuracy, slightly below the 60\% majority-class baseline. In contrast, {\methodname} reaches 71.6\% accuracy across all predictions. At high confidence ($P \geq 0.9$), {\methodname} achieves 93.2\% accuracy, a regime the ensemble cannot access without additional calibration because it produces no per-prediction probability. The 4-model ensemble retains a weak continuous-ranker signal ($\rho = 0.229$) but offers no binary-classification advantage over the majority-class baseline at the median threshold.

In practice, {\methodname} adds less than 1\% computational overhead to a single MLIP forward pass. The classifier has ~567K parameters, two orders of magnitude smaller than either backbone.

\subsection{{\methodname} extends to MACE representations}\label{sec:mace}

\begin{figure}[h]
\centering
\includegraphics[width=\textwidth]{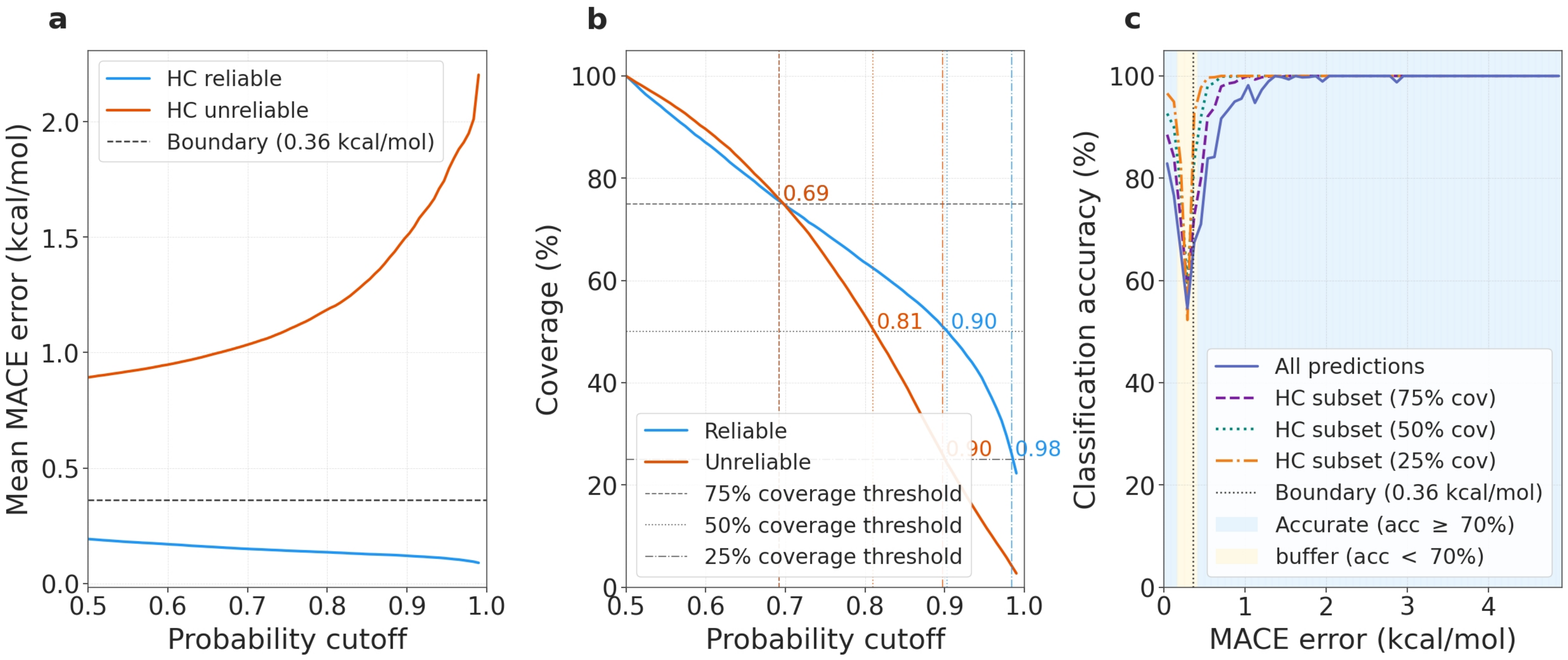}
\caption{\textbf{{\methodname} performance on MACE-OFF23 (50,195 held-out molecules, boundary at 0.36~kcal/mol).}
(a) Mean MACE error as a function of probability cutoff for the reliable (blue) and unreliable (orange) classes.
(b) High-confidence coverage as a function of probability cutoff. Vertical and horizontal markers indicate cutoffs corresponding to 25\%, 50\%, and 75\% coverage thresholds.
(c) Overall classification accuracy (solid navy) and HC subset accuracy (dashed lines, one per coverage threshold) as a function of binned MACE error magnitude. Shading indicates accurate (blue, overall accuracy $\geq 70\%$) and buffer (yellow, overall accuracy $< 70\%$) regions.}\label{fig:mace_results}
\end{figure}

We apply {\methodname} to MACE-OFF23 with no changes to the classifier architecture, hyperparameters, or training protocol used for AIMNet2 in Section~\ref{sec:aimnet2}; the only modification is that per-atom embeddings are read from the invariant ($L=0$) scalar components
of the $\mathbf{B}$-features (Eq.~10 of~\cite{batatia2022mace}) from the final tensor product block. This serves as our primary test of architecture transfer. Additional details about the MACE architecture and training data are provided in Methods (Section \ref{sec:MACE_setup}).

\begin{table}[h]
\caption{\textbf{{\methodname} high-confidence performance for MACE-OFF23 (boundary at 0.36~kcal/mol).} Evaluated on 50,195 held-out molecules. $\bar{e}_\text{rel}$ and $\bar{e}_\text{unrel}$ are mean MACE errors (kcal/mol) for the HC-reliable and HC-unreliable subsets respectively.}\label{tab:mace_hc}
\small
\begin{tabular*}{\columnwidth}{@{\extracolsep\fill}lcccccc}
\toprule
Cutoff & Coverage & Acc. & MCC & F1 & $\bar{e}_\text{rel}$ & $\bar{e}_\text{unrel}$ \\
 & & & & & (kcal/mol) & (kcal/mol) \\
\midrule
All ($P \geq 0.5$) & 100.0\% & 80.5\% & 0.612 & 0.826 & 0.19 & 0.89 \\
$P \geq 0.6$       &  88.5\% & 83.8\% & 0.679 & 0.857 & 0.17 & 0.95 \\
$P \geq 0.7$       &  74.6\% & 87.5\% & 0.752 & 0.889 & 0.15 & 1.03 \\
$P \geq 0.8$       &  57.3\% & 91.5\% & 0.832 & 0.920 & 0.13 & 1.19 \\
$P \geq 0.9$       &  35.5\% & 96.0\% & 0.918 & 0.952 & 0.12 & 1.50 \\
$P \geq 0.95$      &  23.5\% & 97.9\% & 0.952 & 0.967 & 0.11 & 1.77 \\
$P \geq 0.99$      &  10.7\% & 99.6\% & 0.985 & 0.987 & 0.09 & 2.20 \\
\bottomrule
\end{tabular*}
\end{table}

{\methodname} achieves 80.5\% overall accuracy and MCC = 0.612 on the 50,195-molecule test set (Table~\ref{tab:mace_hc}). {\methodname}'s absolute metrics are higher on MACE-OFF23 than on AIMNet2. The two evaluations differ simultaneously in test-distribution overlap, training-set size, class-boundary value, richness of MLIP internal representation, and backbone architecture. We therefore do not attribute the numerical gap to any single factor here, and return to scaling-related implication in Section~\ref{sec:discussion}. The operationally relevant result is that identical {\methodname} hyperparameters transfer across both backbones without any tuning. As with AIMNet2, confidence monotonically tracks error in both directions (Fig.~\ref{fig:mace_results}(a)): mean error for HC-reliable predictions decreases from 0.19~kcal/mol at $P \geq 0.5$ to 0.09~kcal/mol at $P \geq 0.99$, while HC-unreliable predictions increase from 0.89 to 2.20~kcal/mol over the same range. Coverage filtering shows the expected monotonic tradeoff (Fig.~\ref{fig:mace_results}(b)). At $P \geq 0.7$, {\methodname} retains 74.6\% of the test set at 87.5\% accuracy (MCC = 0.752); at $P \geq 0.9$, 35.5\% of molecules are retained at 96.0\% accuracy (MCC = 0.918). Fig.~\ref{fig:mace_results}(c) shows that the buffer region where overall accuracy falls below 70\% is narrow and centered tightly on the 0.36~kcal/mol boundary, and HC subset lines sit well above the overall accuracy curve throughout. Furthermore, MACE-OFF23 shows near-perfect calibration (ECE = 0.011, Brier score = 0.136; SI Fig. S7b), consistent with the rich equivariant representations.

These results confirm that {\methodname} is not contingent on a particular representation family. AIMNet2's AIM embeddings and MACE's scalar node features are constructed from different physical priors and trained on datasets differing by two orders of magnitude in size. Both support {\methodname} with identical classifier hyperparameters. The only backbone-specific change is zeroing the optional partial-charge injection term for MACE, which does not expose atomic charges.

\subsection{Attention scores provide an interpretable atom-level view of prediction reliability}\label{sec:attention}

\begin{figure}[h]
\centering
\includegraphics[width=\textwidth]{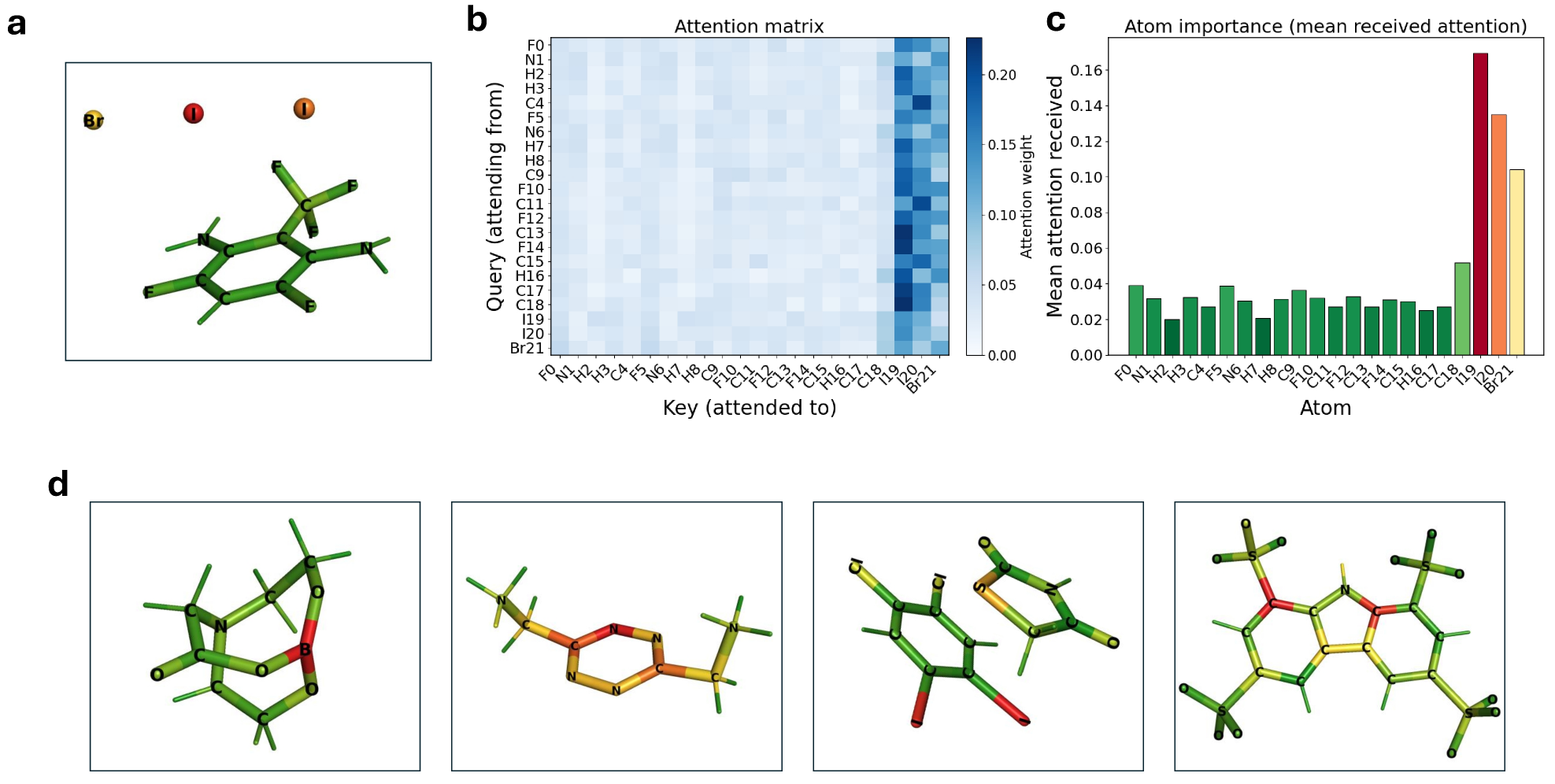}
\caption{\textbf{Atom-level reliability interpretation via {\methodname} attention scores.}
\textbf{(a)} Molecular environment containing iodine and bromine colored by per-atom importance
score $s_{ia}$ (green--yellow--red; red = high importance).
\textbf{(b)} Mean-over-heads self-attention matrix for the molecule in (a); query
atoms along the $y$-axis, key atoms along the $x$-axis.
\textbf{(c)} Per-atom importance score $s_{ia}$; iodine (I) and bromine (Br) dominate.
\textbf{(d)} Four further examples of diverse chemical environments flagged as
unreliable by {\methodname} (all molecules, $P(\text{unreliable}) \geq 0.95$).
We emphasize that this is a qualitative illustration of the attention-score output;
systematic validation that flagged atoms correspond to training-data underrepresentation
would require targeted DFT benchmarks and dataset analysis beyond the scope of this work.
}\label{fig:attention}
\end{figure}

{\methodname}'s multi-head self-attention mechanism yields chemically interpretable,
per-atom importance scores at no additional computational cost.
For each molecule $i$ and attention head $m$, the attention matrix
$A_i^{(m)} \in \mathbb{R}^{N_i \times N_i}$ is computed as described in
Section~\ref{sec:methods}, where entry $A_{i,ab}^{(m)}$ is the weight atom $a$ (query,
$y$-axis in Fig.~\ref{fig:attention}b) places on atom $b$ (key, $x$-axis).
To obtain a scalar importance score for each atom $b$, we sum over all queries,
i.e., we ask how much total attention atom $b$ \emph{receives} from every other atom
(including itself) across all heads:
\begin{equation}
s_{ib} = \frac{1}{Z_i}\sum_{m=1}^{M}\sum_{a=1}^{N_i} A_{i,ab}^{(m)},
\label{eq:importance}
\end{equation}
where $Z_i$ normalizes so that $\sum_b s_{ib} = 1$. Atoms with high $s_{ib}$ are strongly \emph{attended to} by the rest of the molecule,
meaning they exert disproportionate influence on the molecular embedding that drives
the final reliability decision.

Fig.~\ref{fig:attention} illustrates this on high-confidence unreliable predictions from the 3.76M test set for AIMNet2 (Section \ref{sec:aimnet2}).
In panel (a), a molecular environment containing iodine and bromine is shown: both heavy halogens
accumulate disproportionately high importance scores (panels b-c).
This is chemically sensible, since heavy halogens create unusual local electron density
distributions that can lie outside the training distribution of AIMNet2, which {\methodname} correctly identifies.
Panel (d) shows four additional examples, selected to span diverse chemistry, in which
attention mass concentrates on chemically unusual moieties (heavy halogens, hypervalent
sulfur, phosphorus-halogen motifs). These are illustrative rather than a systematic
evaluation; the population-level signal is quantified in Section~\ref{sec:umap}.

In deployment, a practitioner can inspect the importance map of any flagged molecule
to identify \emph{which} structural feature triggered the unreliable prediction.
Aggregating importance scores over a large set of unreliable predictions surfaces
recurrent structural motifs that
the classifier's representation treats as discriminative of unreliability.
This creates a closed feedback loop, {\methodname} not only flags unreliable
predictions but actively guides where to acquire new training data, complementing
the active learning workflow described in Section~\ref{sec:active_learning}.

\subsection{{\methodname} molecular embeddings reveal chemical space coverage and reliability boundaries}\label{sec:umap}

\begin{figure}[t]
\centering
\includegraphics[width=\textwidth]{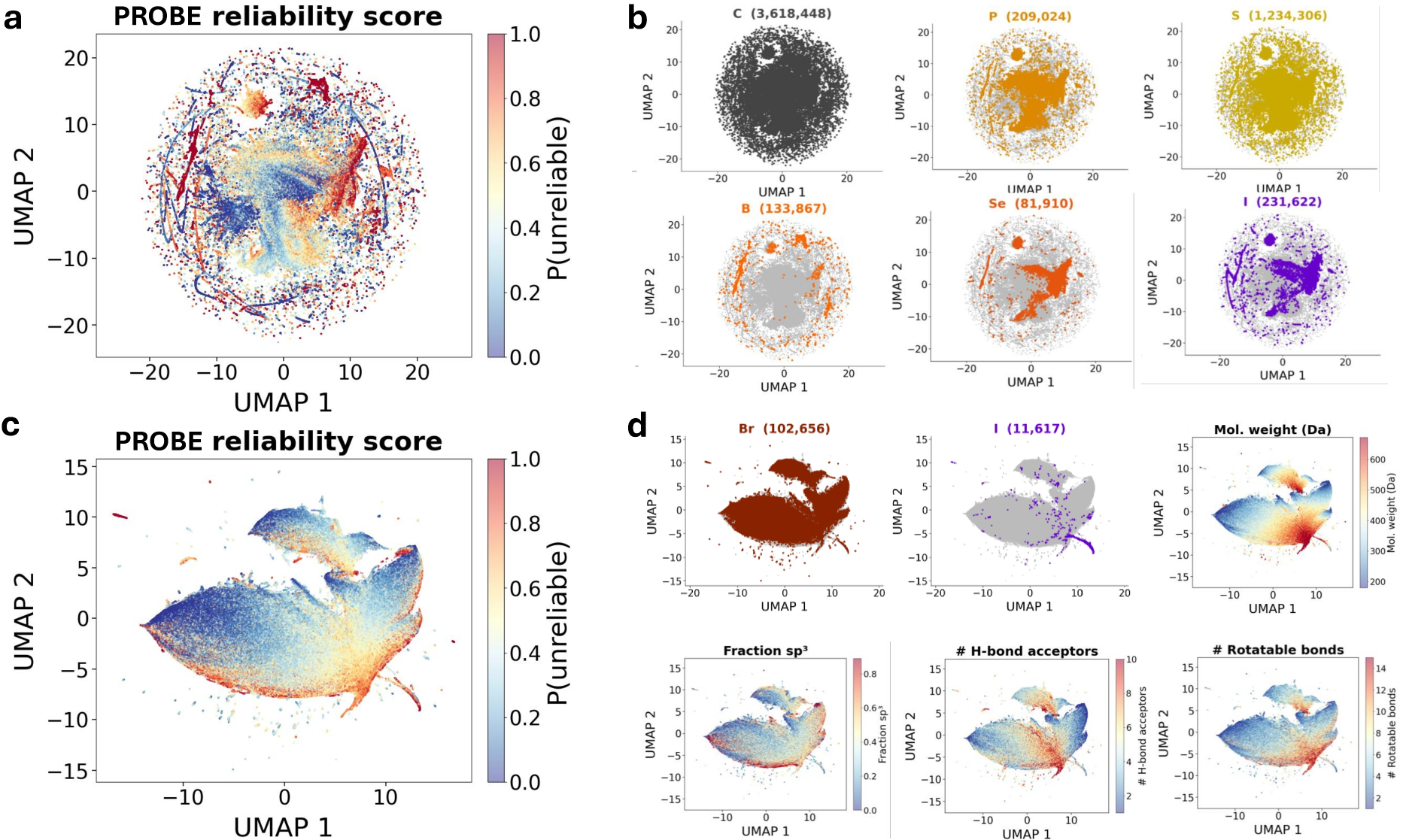}
\caption{\textbf{UMAP projections of {\methodname} molecular embeddings.}
\textbf{(a,b)} 3.76M held-out AIMNet2 test set.
\textbf{(c,d)} 2.4M biologically relevant molecules from Reidenbach et al.
\textbf{(a,c)} $P(\text{unreliable})$ projected onto UMAP coordinates.
\textbf{(b)} Selected element presence overlays (molecule counts in parentheses).
\textbf{(d)} Element presence (Br, I) and RDKit descriptor projections.
}\label{fig:umap}
\end{figure}

The molecular embedding produced by {\methodname} is trained
solely to separate reliable from unreliable MLIP predictions. If this embedding has learned chemically meaningful structure, molecules should organize into spatially distinct reliable and unreliable regions, for a \emph{completely unseen} dataset, and requiring no reference energies at inference time.
We first projected the molecular embeddings of the 3.76M held-out test set created by {\methodname} (input to classifier, Section~\ref{sec:aimnet2}) onto two dimensions using UMAP~\cite{mcinnes2018umap}
and colored each point by $P(\text{unreliable})$ (Fig.~\ref{fig:umap}a).
The embedding visibly organizes test molecules into reliable- and unreliable-enriched regions across local clusters, consistent with the claim that the embedding captures chemically distinct success and failure modes.

Overlaying element presence onto the same UMAP coordinates (Fig.~\ref{fig:umap}b)
is consistent with specific chemistries concentrating in specific regions of the
embedding, though we do not claim causal attribution of reliability to element
identity alone.
Carbon is ubiquitous across the entire manifold, as expected for organic chemistry
datasets.
Sulfur follows a similar but slightly narrower distribution.
By contrast, boron, selenium, and iodine map predominantly onto high-$P(\text{unreliable})$
regions, associating their presence with AIMNet2 failure, while phosphorus occupies a mixed region.
The complete element projection for all 14 AIMNet2 elements is provided in SI
Fig.~S4.

We additionally applied {\methodname}
without retraining to a completely separate 2.4M-molecule set of biologically relevant molecules
published by Reidenbach et al.~\cite{nikitin2025scalable}. {\methodname} labels 82\% of these molecules as reliable ($P(\text{reliable}) \geq 0.5$)
and the remaining 18\% as unreliable; because no reference energies are available here,
this is a classifier self-report rather than a ground-truth measurement.It is consistent in direction with the set's drug-like, CHNO-dominated
composition. The UMAP projection (Fig.~\ref{fig:umap}c) shows a compact, predominantly blue
manifold with a distinct tail of high-$P(\text{unreliable})$ molecules, again
spatially separated from the reliable core.
Element presence and RDKit~\cite{landrum2013rdkit} descriptors computed from connectivity (Fig.~\ref{fig:umap}d)
show that iodine-containing molecules for example, concentrate in the unreliable tail, while bromine
is distributed more broadly.
Among molecular properties, higher molecular weight, TPSA, number of H-bond
acceptors, and rotatable bond count all correlate positively with $P(\text{unreliable})$. Full element and descriptor projections for this dataset are provided in
SI Figs.~S5, S6.

These results demonstrate that the {\methodname} molecular embedding encodes
actionable chemical information about MLIP reliability.
A practitioner can use this analysis on any new compound library to map where
molecules fall relative to known failure modes without ground truth labels, and prioritize structural classes for further validation or targeted data acquisition.

\subsection{Retrospective active learning}\label{sec:active_learning}

As a proof of concept, we show how {\methodname} can guide targeted data acquisition for MLIP retraining. We compare two selection strategies applied to AIMNet2 retraining: \textbf{Ensemble} (molecules ranked by ensemble standard deviation) and \textbf{{\methodname}} (molecules ranked by $P(\text{unreliable})$). Both strategies operate on the same fixed pool of $\sim$20M molecules and are evaluated on a held-out test set of 13.6M molecules.

Cycle~0 trains an ensemble of four independent AIMNet2 models on 1M molecules drawn by stratified random sampling; for {\methodname}-based selection we use a single AIMNet2 model from this four-member Cycle-0 ensemble. Each subsequent cycle augments the training set with 1M additional molecules selected by the respective strategy, with model parameters loaded from the previous cycle. The ensemble Cycle~0 model starts with lower error than the single {\methodname} model (RMSE = 3.030 vs.\ 3.234~kcal/mol), a consequence of ensemble averaging (i.e., better models than the one chosen for {\methodname}) rather than data selection.

\begin{table}[h]
\caption{\textbf{Retrospective active learning: RMSE (kcal/mol) across cycles.} Each cycle adds 1M molecules selected by the respective strategy. Evaluated on a fixed 13.6M molecule held-out test set. $\Delta$RMSE = cumulative change from Cycle~0. The ensemble Cycle~0 baseline is lower due to ensemble averaging, not data selection.}\label{tab:active_learning}
\small
\begin{tabular*}{\columnwidth}{@{\extracolsep\fill}llcc}
\toprule
Cycle & Method & RMSE (kcal/mol)& $\Delta$ (kcal/mol)\\
\midrule
C0 & {\methodname} & 3.234 & --- \\
C0 & Ensemble      & 3.030 & --- \\
\midrule
C1 & {\methodname} & 2.903 & $-$0.331 \\
C1 & Ensemble      & 2.907 & $-$0.123 \\
\midrule
C2 & {\methodname} & 2.709 & $-$0.525 \\
C2 & Ensemble      & 2.817 & $-$0.213 \\
\bottomrule
\end{tabular*}
\end{table}

Across two retrospective active-learning cycles, {\methodname}-guided acquisition reduced RMSE by 0.525~kcal/mol (16.2\%) versus 0.213~kcal/mol (7.0\%) for ensemble-$\sigma$ selection, more than twice the improvement. By Cycle~2, {\methodname} not only closes the initial 0.2~kcal/mol Cycle-0 gap but surpasses the ensemble at roughly one-quarter of the per-cycle retraining compute (one model retrained per cycle versus four). We report this as suggestive rather than definitive: the comparison involves a single seed per strategy, two acquisition cycles, no random-selection control, and simultaneously differs in selection strategy and in number of models retrained. A like-for-like benchmark with matched retraining protocols, across multiple seeds and with a random-selection baseline, is needed to attribute the gap to selection quality alone. We leave this to future work.

\section{Discussion}\label{sec:discussion}

The question of when to trust an MLIP admits two answers: regress the continuous error magnitude, or predict a binary reliability verdict. We argue the second is more tractable: reducing to classification on the backbone's own representations and sufficient for most deployment decisions. Grounding the reliability signal in MLIP internal representations rather than committee disagreement is what drives {\methodname}'s advantage over ensemble-based approaches.

The active-learning advantage {\methodname} demonstrates is the expected consequence of a measured ranking-quality gap. {\methodname} is one of the few single-model UQ methods demonstrated in this work to outperform ensemble-based selection in an MLIP active-learning scenario; prior comparative studies~\cite{tan2023} found that single-model UQ does not consistently match ensembles for MLIP error estimation, and our retrospective experiment suggests the binary-classification framing combined with representation-based scoring can close or reverse this gap in at least one setting, at roughly one-quarter the retraining cost per cycle. The same explanation underlies the static benchmark in Section~\ref{sec:ensemble}: ensemble $\sigma$ delivers a weak correlation against the AIMNet2 error, whereas {\methodname}'s $P(\text{unreliable})$ tracks mean error monotonically across the full $[0.5, 1.0]$ probability range on \emph{both} backbones (Figs.~\ref{fig:aimnet2_results}(b), \ref{fig:mace_results}(a)). A weak ranker produces noisy acquisition decisions; a monotonic one does not.

For foundation-scale MLIPs (UMA~\cite{wood2025uma}, AllScAIP~\cite{qu2026allscaip}), maintaining $N$ independent model copies is impractical. {\methodname}'s post-hoc frozen-backbone design adds negligible inference overhead and requires no backbone retraining, placing it alongside conformal prediction~\cite{ho2025}, shallow ensembles~\cite{kellner2024shallow}, and prediction-rigidity methods~\cite{bigi2024prediction} among the few UQ routes practical at foundation scale. Among these, {\methodname} is the only one that delivers a binary reliability decision \emph{and} per-atom interpretability (Table~\ref{tab:peer_compare}).

\begin{table}[h]
\caption{\textbf{Single-backbone UQ methods for MLIPs compared.} ``Post-hoc'' denotes methods requiring no backbone modification or retraining. $\checkmark$ = yes; $\times$ = no; ``partial'' = additional trained heads but no full backbone retraining. {\methodname} is the only method that is simultaneously single-backbone, post-hoc, binary-output, and per-atom interpretable.}\label{tab:peer_compare}
\small
\begin{tabular*}{\columnwidth}{@{\extracolsep\fill}lccccc}
\toprule
Method & Ref. & Single backbone & Post-hoc & Output type & Per-atom signal \\
\midrule
Shallow ensembles      & \cite{kellner2024shallow}    & $\checkmark$ & partial & continuous        & $\times$ \\
Prediction rigidity    & \cite{bigi2024prediction}    & $\checkmark$ & $\checkmark$ & continuous        & $\times$ \\
Latent-distance UQ     & \cite{musielewicz2024latent} & $\checkmark$ & $\checkmark$ & continuous        & $\checkmark$ \\
Multi-head committee   & \cite{beck2025}              & $\checkmark$ & partial & continuous        & $\times$ \\
Conformal prediction   & \cite{ho2025}                & $\checkmark$ & $\checkmark$ & coverage interval & $\times$ \\
\rowcolor{blue!10}{\methodname} (this work) & ---   & $\checkmark$ & $\checkmark$ & binary probability & $\checkmark$ (attention) \\
\bottomrule
\end{tabular*}
\end{table}

Beyond tractability, the backbone comparison hints at a favorable scaling trend. {\methodname} achieves notably higher discrimination on MACE-OFF23 than on AIMNet2 despite MACE being trained on $20\times$ less data. Since the classifier is identical in both cases, the gain must come from the richer representation MACE exposes. This implies a favorable scaling trajectory: as backbone representations improve, {\methodname}'s reliability signal strengthens, suggesting that next-generation foundation MLIPs (UMA~\cite{wood2025uma}, AllScAIP~\cite{qu2026allscaip}) should yield better, not worse, post-hoc reliability estimates at fixed classifier cost. A systematic scaling study across backbone sizes is a natural next step.

The 50th-percentile boundary yields stable training and a balanced class split. Moving the boundary substantially degrades performance as the classifier defaults to the majority class. Users targeting a specific error tolerance are better served by adjusting the probability cutoff on a balanced model rather than redefining the boundary. Importantly, the boundary's absolute value (1.64~kcal/mol for AIMNet2; 0.36~kcal/mol for MACE-OFF23) is not comparable across backbones; users needing a fixed error tolerance should treat $P(\text{reliable})$ as a relative ranking signal calibrated on a small labeled reference set.

The attention mechanism yields per-atom importance scores at no additional computational cost, identifying exactly which structural motifs drive an unreliable prediction and providing actionable guidance on where additional training data or DFT validation is most needed. Two independent signals converge on the same AIMNet2 failure mode: locally, attention maps assign the highest importance to iodine and bromine in high-confidence unreliable predictions (Fig.\ref{fig:attention}); globally, the UMAP projection of {\methodname}'s molecular embedding concentrates iodine-containing molecules in the unreliable tail on both the 3.76M test set and the Reidenbach 2.4M library (Fig.\ref{fig:umap}b,d). Their agreement across different granularities and datasets supports the interpretation that {\methodname} has surfaced a genuine training-data gap rather than an artefact of either analysis.

Several limitations apply: (i) {\methodname} is a binary classifier, applications requiring continuous uncertainty estimates would need a multi-class or regression extension. (ii) Like most UQ approaches, performance degrades for chemical environments completely absent from training data. (iii) Training requires reference labels for backbone training molecules, which may be unavailable for large foundation models. (iv) Only energy-prediction reliability is evaluated here; extending {\methodname} to per-atom quantities such as forces is conceptually direct but not demonstrated in this work. (v) The reliability boundary depends on the backbone's error distribution and must be recalibrated when the backbone changes. (vi) PROBE's reliability labels are currently derived from DFT-based reference energies and thus cannot capture errors arising from DFT approximations themselves, but this limitation is not fundamental. Following Kellner et al. \cite{2604.24607}, PROBE could be calibrated using experimentally-derived reliability labels wherever such data exist, extending its scope beyond DFT-level errors to empirical measurements.

\textbf{Future directions.} Three extensions are natural: \emph{periodic systems} (applying {\methodname} to materials foundation MLIPs such as MACE-MP~\cite{kovacs2023macemp} and UMA~\cite{wood2025uma}); \emph{force uncertainty} (classifying per-atom force-error percentiles, most relevant to molecular dynamics); and \emph{genuine out-of-distribution evaluation} (training on CHNO, testing on heavy halogens or transition metals). Each is enabled by the post-hoc architecture-agnostic design.

\textbf{Conclusion.} The backbone's own internal representations, fed to a lightweight discriminative classifier, yield a reliability verdict sharper than ensemble standard deviation and cheaper to deploy at foundation scale. This verdict is actionable for the binary decisions that dominate practice: whether to include a molecule in the next active-learning cycle, accept a predicted geometry, or trigger a DFT recalculation. Within its scope, {\methodname} turns ``is this prediction trustworthy?'' into ``with what probability?'', without modifying or retraining the underlying model. As MLIPs scale toward foundation models driving autonomous simulation workflows, knowing when to trust them becomes a prerequisite. {\methodname} addresses this requirement for the binary-decision regime.

\section{Methods}\label{sec:methods}
\subsection{{\methodname} Architecture and Training}\label{sec:arch}
{\methodname} acts as a lightweight binary classifier that takes atom-level representations and other atom-level predictions e.g., partial charges alongside the predicted energy from a single, frozen MLIP as input. We derive training labels from the absolute per-molecule energy error, $\epsilon_m = |\hat{E}^{(m)} - E^{(m)}_\text{ref}|$. The class boundary is itself a percentile of this same training distribution, and the loss is $\sqrt{N}$-normalized. Size-induced error inflation is therefore partially absorbed rather than confounded with unreliability. Per-atom or size-normalized error labels are a straightforward drop-in alternative. Rather than predicting the exact error magnitude, {\methodname} predicts the probability that the error lies below or above a chosen threshold, set as a percentile of the empirical error distribution of the training data. All the experiments in this work use the 50th-percentile boundary, yielding balanced classes.

\textbf{Atom Encoder.} Given a molecular configuration, the backbone MLIP $\mathcal{M}_\theta$ performs a standard forward pass, producing per-atom internal embeddings $\mathbf{h}_i \in \mathbb{R}^{d}$ alongside the predicted energy $\hat{E}$ and atom-level quantities such as partial charges $q_i$. The atom encoder $\text{MLP}_\text{enc} : \mathbb{R}^d \to \mathbb{R}^{256}$ is a three-layer network with hidden dimensions $(d, 256, 128, 256)$ with LayerNorm, GELU activations, and dropout ($p=0.1$). For AIMNet2, $\mathbf{h}_i$ is the per-atom AIM vector; for MACE, it is the invariant ($L{=}0$) scalar components
of the $\mathbf{B}$-features from the final product block. Invariant features are the natural input to a classifier whose target (absolute energy error) is itself a scalar invariant, and using them preserves compatibility without imposing equivariance constraints on the {\methodname} head. These embeddings are passed through the atom encoder: $\mathbf{z}_i = \text{MLP}_\text{enc}(\mathbf{h}_i)$, producing atom-level features relevant for the classification task. Partial charges are incorporated via a learned linear projection: $\mathbf{z}_i \leftarrow \mathbf{z}_i + \mathbf{W}_q q_i$.

\textbf{Molecule encoder.} Atom features are processed through a multi-head self-attention layer with $H$ heads:
\begin{equation}
\mathbf{z}'_i = \mathbf{z}_i + \text{LayerNorm}\left(\sum_{h=1}^{H} \sum_{j=1}^{N} \alpha^{(h)}_{ij} \mathbf{V}^{(h)} \mathbf{z}_j\right)
\end{equation}
where $\alpha^{(h)}_{ij} = \text{softmax}_j\left(\mathbf{z}_i^\top \mathbf{Q}^{(h)\top} \mathbf{K}^{(h)} \mathbf{z}_j / \sqrt{d_k}\right)$ are the attention weights, masked to exclude padding atoms. Each atom aggregates context from all other atoms in the molecule, propagating global chemical environment information into the per-atom representation. Multi-head self-attention operates with $H=32$ heads of dimension $d_k = 8$, giving a total attention dimension of 256. The molecular descriptor $\mathbf{g} \in \mathbb{R}^{514}$ is formed by concatenating masked mean-pool and max-pool of the attended atom features (each $\mathbb{R}^{256}$), the scalar predicted energy $\hat{E}$, and atom count $N$:
\begin{equation}
\mathbf{g} = \left[\frac{1}{N}\sum_i \mathbf{z}'_i \;\Big\|\; \max_i \mathbf{z}'_i \;\Big\|\; \hat{E} \;\Big\|\; N\right] 
\end{equation}

This descriptor is projected to a 256-dimensional molecular embedding $\mathbf{e} = \mathbf{W}_\text{proj}\,\mathbf{g} \in \mathbb{R}^{256}$ via a learned linear map $\mathbf{W}_\text{proj} \in \mathbb{R}^{256 \times 514}$. The embedding $\mathbf{e}$ serves as a general-purpose molecular representation and is used directly for chemical space coverage analysis.

\textbf{Classifier.} The classifier $\text{MLP}_\text{cls} : \Delta^1$ then maps $\mathbf{e}$ to binary reliability probabilities. The class boundary is set at the $p$th percentile of the training error distribution (default $p = 50$, yielding balanced classes). The loss function is cross-entropy normalized by $\sqrt{N_m}$ to prevent large molecules from dominating:
\begin{equation}
\mathcal{L} = \frac{1}{|\mathcal{B}|} \sum_{m \in \mathcal{B}} \frac{w_{y_m} \cdot \text{CE}(f_\phi(\mathbf{e}_m), y_m)}{\sqrt{N_m}}, \qquad w_c = \frac{|\mathcal{D}|}{2\,|\mathcal{D}_c|}
\end{equation}
where $y_m \in \{0, 1\}$ is the binary reliability label and $|\mathcal{D}_c|$ is the number of training samples in class $c$. When classes are balanced, $w_c = 1$ and the loss reduces to plain normalized cross-entropy. The classifier $\text{MLP}_\text{cls} : \mathbb{R}^{256} \to \Delta^1$ maps $\mathbf{e}$ to reliability probabilities through hidden dimensions $[128, 32]$ with LayerNorm, GELU activations, and dropout ($p=0.1$). The full model has \~567K trainable parameters.

\textbf{Optimization.} Parameters $\phi$ are optimized with AdamW (learning rate $\eta = 5 \times 10^{-5}$, weight decay $\lambda = 10^{-4}$, gradient clipping with max norm 1.0). A ReduceLROnPlateau scheduler reduces $\eta$ by factor 0.9 after 5 epochs without validation loss improvement, with a minimum learning rate of $5 \times 10^{-6}$. Early stopping is applied with patience 25 epochs. All models use a 90/10 train/validation split; a fully held-out test set is used exclusively for final evaluation. The MLIP backbone is fully frozen throughout.

\subsection{Problem Formulation}\label{sec:problem}

\begin{algorithm}[t]
\caption{{\methodname} Training}\label{alg:rumi}
\begin{algorithmic}[1]
\Require Frozen MLIP $\mathcal{M}_\theta$; calibration set $\mathcal{D}$; percentile $p \in (0,1)$; max epochs $T$
\Ensure {\methodname} classifier parameters $\phi$
\State \textbf{Label generation:}
\State \quad Compute $\epsilon_m \leftarrow |\mathcal{M}_\theta(\mathbf{R}^{(m)},\mathbf{Z}^{(m)}) - E^{(m)}_\text{ref}|$ for all $m \in \mathcal{D}$
\State \quad Set boundary $b_p \leftarrow \mathrm{Quantile}_p(\{\epsilon_m\})$; assign $y_m \leftarrow \mathbf{1}[\epsilon_m \geq b_p]$
\State \quad Compute class weights $w_c \leftarrow |\mathcal{D}|\,/\,(2\,|\mathcal{D}_c|)$, where $\mathcal{D}_c = \{m : y_m = c\}$
\For{epoch $t = 1, \ldots, T$}
    \For{each mini-batch $\mathcal{B} \subseteq \mathcal{D}$}
        \State \textbf{MLIP representation extraction (no gradient):}
        \State \quad $\{\mathbf{h}_i^{(m)}\},\, \hat{E}^{(m)},\, \{q_i^{(m)}\} \leftarrow \mathcal{M}_\theta(\mathbf{R}^{(m)},\mathbf{Z}^{(m)})$ \quad $\forall m \in \mathcal{B}$
        \State \textbf{Atom encoding:}
        \State \quad $\mathbf{z}_i^{(m)} \leftarrow \mathrm{MLP}_\mathrm{enc}(\mathbf{h}_i^{(m)}) + \mathbf{W}_q\,q_i^{(m)}$
        \State \textbf{Multi-head self-attention with masking:}
        \State \quad ${\mathbf{z}'}^{(m)} \leftarrow \mathbf{z}^{(m)} + \mathrm{LayerNorm}\!\left(\mathrm{MHSA}\!\left(\mathbf{z}^{(m)},\,\mathrm{mask}^{(m)}\right)\right)$
        \State \textbf{Molecular pooling:}
        \State \quad $\mathbf{g}^{(m)} \leftarrow \Bigl[\tfrac{1}{N^{(m)}}\textstyle\sum_i \mathbf{z}'^{(m)}_i \;\Big\|\; \max_i \mathbf{z}'^{(m)}_i \;\Big\|\; \hat{E}^{(m)} \;\Big\|\; N^{(m)}\Bigr]$
        \State \textbf{Molecular embedding:}
        \State \quad $\mathbf{e}^{(m)} \leftarrow \mathbf{W}_\text{proj}\,\mathbf{g}^{(m)}$
        \State \textbf{Classification and loss:}
        \State \quad $\hat{\mathbf{y}}^{(m)} \leftarrow \mathrm{softmax}\!\left(\mathrm{MLP}_\mathrm{clf}(\mathbf{e}^{(m)})\right)$
        \State \quad $\mathcal{L} \leftarrow \dfrac{1}{|\mathcal{B}|}\displaystyle\sum_{m \in \mathcal{B}} \dfrac{w_{y_m}\cdot\mathrm{CE}(\hat{\mathbf{y}}^{(m)},\,y_m)}{\sqrt{N^{(m)}}}$
        \State Update $\phi \leftarrow \phi - \eta\,\nabla_\phi\,\mathcal{L}$ \quad (AdamW)
    \EndFor
    \State Reduce $\eta$ on validation loss plateau; apply early stopping
\EndFor
\end{algorithmic}
\end{algorithm}

Let $\mathcal{M}_\theta$ denote a trained MLIP with frozen parameters $\theta$, mapping a molecular configuration $(\mathbf{R}, \mathbf{Z}) \in \mathbb{R}^{N \times 3} \times \mathbb{Z}^N$ to a predicted energy $\hat{E} \in \mathbb{R}$ and per-atom latent representations $\{\mathbf{h}_i\}_{i=1}^{N} \subset \mathbb{R}^d$. Given a labeled calibration dataset $\mathcal{D} = \{(\mathbf{R}^{(m)}, \mathbf{Z}^{(m)}, E^{(m)}_\text{ref})\}_{m=1}^{|\mathcal{D}|}$ with reference energies $E^{(m)}_\text{ref}$, define the per-molecule absolute error
\begin{equation}
    \epsilon_m = \bigl|\hat{E}^{(m)} - E^{(m)}_\text{ref}\bigr|.
\end{equation}

Let $b_p = \mathrm{Quantile}_p\!\left(\{\epsilon_m\}_{m=1}^{|\mathcal{D}|}\right)$ denote the $p$th percentile of the empirical error distribution. We assign binary reliability labels:
\begin{equation}
    y_m = \begin{cases} 0 & \text{if } \epsilon_m < b_p \quad \text{(reliable)} \\ 1 & \text{if } \epsilon_m \geq b_p \quad \text{(unreliable).} \end{cases}
\end{equation}

\textbf{Goal.} Learn a classifier $f_\phi : \bigl(\{\mathbf{h}_i\}_{i=1}^N,\, \hat{E},\, N\bigr) \to \Delta^1$ with parameters $\phi$, operating solely on quantities produced by $\mathcal{M}_\theta$ (no access to $E_\text{ref}$ at inference time), such that the predicted probability $f_\phi(\cdot)_1 \approx P(y = 1 \mid \mathbf{R}, \mathbf{Z})$ generalizes to held-out molecules. The parameters $\theta$ of $\mathcal{M}_\theta$ are \emph{never updated}.

Training minimizes a size-normalized, class-weighted cross-entropy:
\begin{equation}
    \mathcal{L} = \frac{1}{|\mathcal{B}|} \sum_{m \in \mathcal{B}} \frac{w_{y_m} \cdot \text{CE}(f_\phi(\mathbf{c}_m),\, y_m)}{\sqrt{N_m}}, \qquad w_c = \frac{|\mathcal{D}|}{2\,|\mathcal{D}_c|}
\end{equation}
where $\mathcal{B}$ is a mini-batch, $N_m$ is the atom count of molecule $m$, and $w_c$ are inverse-frequency class weights. The $\sqrt{N_m}$ normalization prevents large molecules from dominating the gradient. When classes are balanced, $w_c = 1$ and the loss reduces to plain normalized cross-entropy.

This formulation reduces MLIP reliability estimation to supervised binary classification on a learned representation space, avoiding direct regression of the heavy-tailed distribution $p(\epsilon \mid \mathbf{R}, \mathbf{Z})$. The full training procedure is summarized in Algorithm~\ref{alg:rumi}.

\subsection{AIMNet2 training and evaluation}\label{sec:aimnet2_setup}

AIMNet2 is a neural network potential based on the atoms-in-molecules (AIM) framework trained on 20 million hybrid DFT calculations (B97-3c/def2-mTZVP) covering neutral and charged organic molecules across 14 element types. AIMNet2 is initialized from Behler-Parrinello type atomic environment vectors~\cite{behler2007}. In AIMNet2, the per-atom learned representation, the atoms-in-molecules (AIM) vector, is explicitly designed as a general-purpose chemical descriptor functionally analogous to the electron density $\rho(r)$ in DFT. It serves as a common substrate refined through iterative self-consistency, specifically message passing with neural charge equilibration, from which multiple physical observables such as energy, forces, partial charges are projected out through distinct learned readout heads. This multi-task design implies that the representation already encodes rich chemical context.

We trained an ensemble of four AIMNet2 models using the default architecture and training procedure outlined in ~\cite{anstine2025aimnet2} that were fully frozen during {\methodname} training. {\methodname} was trained using the same dataset used to train AIMNet2 models with a 90/10 train/validation split. The 50th-percentile class boundary on the training set is 1.64~kcal/mol, yielding balanced classes (50\%/50\%); the full training-set error distribution is shown in the Supplementary Information (SI) Fig. S1. Applying this fixed 1.64~kcal/mol class boundary (the 50th percentile of the \emph{training} error distribution) to the 3.76M test set yields a 60/40 reliable/unreliable split. 

\subsection{MACE training and evaluation}\label{sec:MACE_setup}

MACE-OFF23~\cite{batatia2022mace,kovacs2023macemp} is a higher-order $E(3)$-equivariant message-passing neural network that achieves body-order completeness with only two interaction layers. Each interaction layer constructs equivariant features up to angular momentum $L$. MACE-OFF23 is a transferable organic force field trained on the SPICE dataset of drug-like organic molecules. We use the publicly available large model and its training set (855,905 molecules, split 90/10 for {\methodname} training and validation). Evaluation uses the original held-out test set of 50,195 molecules. The class boundary is set at 0.36~kcal/mol (50th percentile, balanced classes). The resulting class split is 47.3\% reliable / 52.7\% unreliable on the test set, with mean errors of 0.139 and 1.029~kcal/mol, respectively. Training-set error distribution and train/validation loss curves for training {\methodname} are provided in the SI Fig. S3. {\methodname} accesses the scalar node features (\texttt{node\_feats}) from MACE's final interaction layer as per-atom embeddings, with the MACE backbone cast to float32 and fully frozen. MACE-OFF23 does not expose atomic partial charges in the same manner as AIMNet2; we therefore set $q_i = 0$ in the partial-charge injection term, reducing the atom encoder to $\mathbf{z}_i = \mathrm{MLP}_\mathrm{enc}(\mathbf{h}_i)$ for this backbone. MACE-level performance remaining competitive with AIMNet2 indicates that partial charges are not essential to {\methodname}'s signal when the backbone's per-atom representation is sufficiently expressive.

\section*{Acknowledgements}

S.M. would like to thank Filipp Gusev, Bhupalee Kalita, and Roman Zuabatyuk for insightful discussions. This work was made possible by the Office of Naval Research (ONR) through support provided by the Energetic Materials Program (MURI grant no.\ N00014-21-1-2476). The authors acknowledge computing resources provided through the Advanced Cyberinfrastructure Coordination Ecosystem: Services and Support (ACCESS) program, specifically the NCSA Delta and Delta AI systems. This work was performed, in part, at the Center for Integrated Nanotechnologies, an Office of Science User Facility operated for the U.S.\ Department of Energy (DOE) Office of Science by Los Alamos National Laboratory (Contract 89233218CNA000001) and Sandia National Laboratories (Contract DE-NA-0003525). Claude Code (Opus 4.5 / Sonnet 4.5) was employed in the implementation of this work.

\section*{Data Availability}

AIMNet2 training data is available through the AIMNet2 repository. MACE training data (SPICE dataset) is publicly available. {\methodname} model weights and inference code will be made available at publication.

\section*{Code Availability}

{\methodname} training and inference code, along with analysis notebooks, will be released at \url{https://github.com/isayevlab/\methodname} upon publication.

\section*{Author Contributions}

S.M. and O.I. designed research; S.M. developed the framework for {\methodname}; S.M. and I.C. performed research; S.M., I.C., O.I. wrote the paper.

\section*{Competing Interests}

The authors declare no competing interests.

\bibliography{references}

@article{behler2007,
  title={Generalized neural-network representation of high-dimensional potential-energy surfaces},
  author={Behler, J{\"o}rg and Parrinello, Michele},
  journal={Physical review letters},
  volume={98},
  number={14},
  pages={146401},
  year={2007},
  publisher={APS}
}

@article{schutt2018,
  title={Schnet--a deep learning architecture for molecules and materials},
  author={Sch{\"u}tt, Kristof T and Sauceda, Huziel E and Kindermans, P-J and Tkatchenko, Alexandre and M{\"u}ller, K-R},
  journal={The Journal of chemical physics},
  volume={148},
  number={24},
  year={2018},
  publisher={AIP Publishing}
}

@article{batatia2022mace,
  title={MACE: Higher order equivariant message passing neural networks for fast and accurate force fields},
  author={Batatia, Ilyes and Kovacs, David P and Simm, Gregor and Ortner, Christoph and Cs{\'a}nyi, G{\'a}bor},
  journal={Advances in neural information processing systems},
  volume={35},
  pages={11423--11436},
  year={2022}
}

@article{anstine2025aimnet2,
  title={AIMNet2: a neural network potential to meet your neutral, charged, organic, and elemental-organic needs},
  author={Anstine, Dylan M and Zubatyuk, Roman and Isayev, Olexandr},
  journal={Chemical Science},
  volume={16},
  number={23},
  pages={10228--10244},
  year={2025},
  publisher={Royal Society of Chemistry}
}

@article{lakshminarayanan2017,
  title={Simple and scalable predictive uncertainty estimation using deep ensembles},
  author={Lakshminarayanan, Balaji and Pritzel, Alexander and Blundell, Charles},
  journal={Advances in neural information processing systems},
  volume={30},
  year={2017}
}

@article{gomez2018automatic,
  title={Automatic chemical design using a data-driven continuous representation of molecules},
  author={G{\'o}mez-Bombarelli, Rafael and Wei, Jennifer N and Duvenaud, David and Hern{\'a}ndez-Lobato, Jos{\'e} Miguel and S{\'a}nchez-Lengeling, Benjam{\'\i}n and Sheberla, Dennis and Aguilera-Iparraguirre, Jorge and Hirzel, Timothy D and Adams, Ryan P and Aspuru-Guzik, Al{\'a}n},
  journal={ACS central science},
  volume={4},
  number={2},
  pages={268--276},
  year={2018},
  publisher={ACS Publications}
}

@article{schran2020,
  title={Committee neural network potentials control generalization errors and enable active learning},
  author={Schran, Christoph and Brezina, Krystof and Marsalek, Ondrej},
  journal={The Journal of Chemical Physics},
  volume={153},
  number={10},
  year={2020},
  publisher={AIP Publishing}
}

@article{smith2018,
  title={Less is more: Sampling chemical space with active learning},
  author={Smith, Justin S and Nebgen, Ben and Lubbers, Nicholas and Isayev, Olexandr and Roitberg, Adrian E},
  journal={The Journal of chemical physics},
  volume={148},
  number={24},
  year={2018},
  publisher={AIP Publishing}
}

@article{gal2016,
  title={Dropout as a bayesian approximation: Representing model uncertainty in deep learning},
  author={Gal, Yarin and Ghahramani, Zoubin},
  booktitle={international conference on machine learning},
  pages={1050--1059},
  year={2016},
  organization={PMLR}
}

@article{wen2020,
  title={Uncertainty quantification in molecular simulations with dropout neural network potentials},
  author={Wen, Mingjian and Tadmor, Ellad B},
  journal={npj computational materials},
  volume={6},
  number={1},
  pages={124},
  year={2020},
  publisher={Nature Publishing Group UK London}
}

@article{vandermause2020,
  title={On-the-fly active learning of interpretable Bayesian force fields for atomistic rare events},
  author={Vandermause, Jonathan and Torrisi, Steven B and Batzner, Simon and Xie, Yu and Sun, Lixin and Kolpak, Alexie M and Kozinsky, Boris},
  journal={npj Computational Materials},
  volume={6},
  number={1},
  pages={20},
  year={2020},
  publisher={Nature Publishing Group UK London}
}

@article{tan2023,
  title={Single-model uncertainty quantification in neural network potentials does not consistently outperform model ensembles},
  author={Tan, Aik Rui and Urata, Shingo and Goldman, Samuel and Dietschreit, Johannes CB and G{\'o}mez-Bombarelli, Rafael},
  journal={npj Computational Materials},
  volume={9},
  number={1},
  pages={225},
  year={2023},
  publisher={Nature Publishing Group UK London}
}

@book{williams2006gaussian,
  title={Gaussian processes for machine learning},
  author={Williams, Christopher KI and Rasmussen, Carl Edward},
  volume={2},
  number={3},
  year={2006},
  publisher={MIT press Cambridge, MA}
}

@article{kovacs2023macemp,
  title={Mace-off: Short-range transferable machine learning force fields for organic molecules},
  author={Kov{\'a}cs, D{\'a}vid P{\'e}ter and Moore, J Harry and Browning, Nicholas J and Batatia, Ilyes and Horton, Joshua T and Pu, Yixuan and Kapil, Venkat and Witt, William C and Magdau, Ioan-Bogdan and Cole, Daniel J and others},
  journal={Journal of the American Chemical Society},
  volume={147},
  number={21},
  pages={17598--17611},
  year={2025},
  publisher={ACS Publications}
}

@article{hohenberg1964,
  title={Inhomogeneous electron gas},
  author={Hohenberg, Pierre and Kohn, Walter},
  journal={Physical review},
  volume={136},
  number={3B},
  pages={B864},
  year={1964},
  publisher={APS}
}

@article{kohn1965,
  title={Self-consistent equations including exchange and correlation effects},
  author={Kohn, Walter and Sham, Lu Jeu},
  journal={Physical review},
  volume={140},
  number={4A},
  pages={A1133},
  year={1965},
  publisher={APS}
}

@article{batzner2022nequip,
  title={E (3)-equivariant graph neural networks for data-efficient and accurate interatomic potentials},
  author={Batzner, Simon and Musaelian, Albert and Sun, Lixin and Geiger, Mario and Mailoa, Jonathan P and Kornbluth, Mordechai and Molinari, Nicola and Smidt, Tess E and Kozinsky, Boris},
  journal={Nature communications},
  volume={13},
  number={1},
  pages={2453},
  year={2022},
  publisher={Nature Publishing Group UK London}
}

@article{deng2023chgnet,
  title={CHGNet as a pretrained universal neural network potential for charge-informed atomistic modelling},
  author={Deng, Bowen and Zhong, Peichen and Jun, KyuJung and Riebesell, Janosh and Han, Kevin and Bartel, Christopher J and Ceder, Gerbrand},
  journal={Nature Machine Intelligence},
  volume={5},
  number={9},
  pages={1031--1041},
  year={2023},
  publisher={Nature Publishing Group UK London}
}

@article{merchant2023gnome,
  title={Scaling deep learning for materials discovery},
  author={Merchant, Amil and Batzner, Simon and Schoenholz, Samuel S and Aykol, Muratahan and Cheon, Gowoon and Cubuk, Ekin Dogus},
  journal={Nature},
  volume={624},
  number={7990},
  pages={80--85},
  year={2023},
  publisher={Nature Publishing Group UK London}
}

@article{wood2025uma,
  title={Uma: A family of universal models for atoms},
  author={Wood, Brandon M and Dzamba, Misko and Fu, Xiang and Gao, Meng and Shuaibi, Muhammed and Barroso-Luque, Luis and Abdelmaqsoud, Kareem and Gharakhanyan, Vahe and Kitchin, John R and Levine, Daniel S and others},
  journal={arXiv preprint arXiv:2506.23971},
  year={2025}
}

@article{liao2024equiformerv2,
  title={Equiformerv2: Improved equivariant transformer for scaling to higher-degree representations},
  author={Liao, Yi-Lun and Wood, Brandon and Das, Abhishek and Smidt, Tess},
  journal={arXiv preprint arXiv:2306.12059},
  year={2023}
}

@article{qu2026allscaip,
  title={A recipe for scalable attention-based MLIPs: unlocking long-range accuracy with all-to-all node attention},
  author={Qu, Eric and Wood, Brandon M and Krishnapriyan, Aditi S and Ulissi, Zachary W},
  journal={arXiv preprint arXiv:2603.06567},
  year={2026}
}

@article{heid2023,
  title={Characterizing uncertainty in machine learning for chemistry},
  author={Heid, Esther and McGill, Charles J and Vermeire, Florence H and Green, William H},
  journal={Journal of Chemical Information and Modeling},
  volume={63},
  number={13},
  pages={4012--4029},
  year={2023},
  publisher={ACS Publications}
}

@article{kulik2022,
  title={Roadmap on machine learning in electronic structure},
  author={Kulik, Heather J and Hammerschmidt, Thomas and Schmidt, Jonathan and Botti, Silvana and Marques, Miguel AL and Boley, Mario and Scheffler, Matthias and Todorovi{\'c}, Milica and Rinke, Patrick and Oses, Corey and others},
  journal={Electronic Structure},
  volume={4},
  number={2},
  pages={023004},
  year={2022},
  publisher={IOP Publishing}
}

@article{hullermeier2021,
  title={Aleatoric and epistemic uncertainty in machine learning: An introduction to concepts and methods},
  author={H{\"u}llermeier, Eyke and Waegeman, Willem},
  journal={Machine learning},
  volume={110},
  number={3},
  pages={457--506},
  year={2021},
  publisher={Springer}
}

@article{farris2025,
  title={Bayesian Neural Networks versus deep ensembles for uncertainty quantification in machine learning interatomic potentials},
  author={Farris, Riccardo and Telari, Emanuele and Artrith, Nongnuch and Neyman, Konstantin and Bruix, Albert},
  journal={arXiv preprint arXiv:2509.19180},
  year={2025}
}

@article{coscia2025blips,
  title={BLIPs: Bayesian Learned Interatomic Potentials},
  author={Coscia, Dario and de Haan, Pim and Welling, Max},
  journal={arXiv preprint arXiv:2508.14022},
  year={2025}
}

@article{ho2025,
  title={Flexible uncertainty calibration for machine-learned interatomic potentials},
  author={Ho, Cheuk Hin and Ortner, Christoph and Wang, Yangshuai},
  journal={arXiv preprint arXiv:2510.00721},
  year={2025}
}

@article{beck2025,
  title={Multi-head committees enable direct uncertainty prediction for atomistic foundation models},
  author={Beck, Hubert and Simko, Pavol and Schaaf, Lars L and Marsalek, Ondrej and Schran, Christoph},
  journal={The Journal of Chemical Physics},
  volume={163},
  number={23},
  year={2025},
  publisher={AIP Publishing}
}

@article{podryabinkin2017,
  title={Active learning of linearly parametrized interatomic potentials},
  author={Podryabinkin, Evgeny V and Shapeev, Alexander V},
  journal={Computational Materials Science},
  volume={140},
  pages={171--180},
  year={2017},
  publisher={Elsevier}
}

@article{kulichenko2023,
  title={Uncertainty-driven dynamics for active learning of interatomic potentials},
  author={Kulichenko, Maksim and Barros, Kipton and Lubbers, Nicholas and Li, Ying Wai and Messerly, Richard and Tretiak, Sergei and Smith, Justin S and Nebgen, Benjamin},
  journal={Nature computational science},
  volume={3},
  number={3},
  pages={230--239},
  year={2023},
  publisher={Nature Publishing Group US New York}
}

@article{vdoord2023hal,
  title={Hyperactive learning for data-driven interatomic potentials},
  author={van der Oord, Cas and Sachs, Matthias and Kov{\'a}cs, D{\'a}vid P{\'e}ter and Ortner, Christoph and Cs{\'a}nyi, G{\'a}bor},
  journal={npj Computational Materials},
  volume={9},
  number={1},
  pages={168},
  year={2023},
  publisher={Nature Publishing Group UK London}
}

@article{zaverkin2024ubmd,
  title={Uncertainty-biased molecular dynamics for learning uniformly accurate interatomic potentials},
  author={Zaverkin, Viktor and Holzm{\"u}ller, David and Christiansen, Henrik and Errica, Federico and Alesiani, Francesco and Takamoto, Makoto and Niepert, Mathias and K{\"a}stner, Johannes},
  journal={npj Computational Materials},
  volume={10},
  number={1},
  pages={83},
  year={2024},
  publisher={Nature Publishing Group UK London}
}

@article{lu2023equivariant,
  title={On the uncertainty estimates of equivariant-neural-network-ensembles interatomic potentials},
  author={Lu, Shuaihua and Ghiringhelli, Luca M and Carbogno, Christian and Wang, Jinlan and Scheffler, Matthias},
  journal={arXiv preprint arXiv:2309.00195},
  year={2023}
}

@article{bartok2022gpr,
  title={Improved uncertainty quantification for Gaussian process regression based interatomic potentials},
  author={Bart{\'o}k, Albert P and Kermode, James and others},
  journal={arXiv preprint arXiv:2206.08744},
  year={2022}
}

@article{xu2024evidential,
  title={Evidential deep learning for interatomic potentials},
  author={Xu, Han and Cui, Taoyong and Tang, Chenyu and Ma, Jinzhe and Zhou, Dongzhan and Li, Yuqiang and Gao, Xiang and Gong, Xingao and Ouyang, Wanli and Zhang, Shufei and others},
  journal={Nature Communications},
  year={2025},
  publisher={Nature Publishing Group UK London}
}

@article{vita2024ltau,
  title={LTAU-FF: loss trajectory analysis for uncertainty in atomistic force fields},
  author={Vita, Joshua A and Samanta, Amit and Zhou, Fei and Lordi, Vincenzo},
  journal={Machine Learning: Science and Technology},
  volume={6},
  number={1},
  pages={015048},
  year={2025},
  publisher={IOP Publishing}
}

@article{vandermause2020gpr,
  title={Active learning of reactive Bayesian force fields applied to heterogeneous catalysis dynamics of H/Pt},
  author={Vandermause, Jonathan and Xie, Yu and Lim, Jin Soo and Owen, Cameron J and Kozinsky, Boris},
  journal={Nature Communications},
  volume={13},
  number={1},
  pages={5183},
  year={2022},
  publisher={Nature Publishing Group UK London}
}

@article{mcinnes2018umap,
  title={Umap: Uniform manifold approximation and projection for dimension reduction},
  author={McInnes, Leland and Healy, John and Melville, James},
  journal={arXiv preprint arXiv:1802.03426},
  year={2018}
}

@article{nikitin2025scalable,
  title={Applications of modular co-design for de novo 3d molecule generation},
  author={Reidenbach, Danny and Nikitin, Filipp and Isayev, Olexandr and Paliwal, Saee Gopal},
  journal={Digital Discovery},
  volume={5},
  number={2},
  pages={754--768},
  year={2026},
  publisher={Royal Society of Chemistry}
}

@article{mehdi2024enhanced,
  title={Enhanced sampling with machine learning},
  author={Mehdi, Shams and Smith, Zachary and Herron, Lukas and Zou, Ziyue and Tiwary, Pratyush},
  journal={Annual Review of Physical Chemistry},
  volume={75},
  number={1},
  pages={347--370},
  year={2024},
  publisher={Annual Reviews}
}

@article{elyaniv2010selective,
  title={On the Foundations of Noise-free Selective Classification.},
  author={El-Yaniv, Ran and others},
  journal={Journal of Machine Learning Research},
  volume={11},
  number={5},
  year={2010}
}

@article{geifman2017selective,
  title={Selective classification for deep neural networks},
  author={Geifman, Yonatan and El-Yaniv, Ran},
  journal={Advances in neural information processing systems},
  volume={30},
  year={2017}
}

@article{geifman2019selectivenet,
  title={Selectivenet: A deep neural network with an integrated reject option},
  author={Geifman, Yonatan and El-Yaniv, Ran},
  booktitle={International conference on machine learning},
  pages={2151--2159},
  year={2019},
  organization={PMLR}
}

@article{devries2018confidence,
  title={Learning confidence for out-of-distribution detection in neural networks},
  author={DeVries, Terrance and Taylor, Graham W},
  journal={arXiv preprint arXiv:1802.04865},
  year={2018}
}

@article{hendrycks2017baseline,
  title={A baseline for detecting misclassified and out-of-distribution examples in neural networks},
  author={Hendrycks, Dan and Gimpel, Kevin},
  journal={arXiv preprint arXiv:1610.02136},
  year={2016}
}

@article{lee2018mahalanobis,
  title={A simple unified framework for detecting out-of-distribution samples and adversarial attacks},
  author={Lee, Kimin and Lee, Kibok and Lee, Honglak and Shin, Jinwoo},
  journal={Advances in neural information processing systems},
  volume={31},
  year={2018}
}

@article{sun2022knnood,
  title={Out-of-distribution detection with deep nearest neighbors},
  author={Sun, Yiyou and Ming, Yifei and Zhu, Xiaojin and Li, Yixuan},
  booktitle={International conference on machine learning},
  pages={20827--20840},
  year={2022},
  organization={PMLR}
}

@article{yang2024oodsurvey,
  title={Generalized out-of-distribution detection: A survey},
  author={Yang, Jingkang and Zhou, Kaiyang and Li, Yixuan and Liu, Ziwei},
  journal={International Journal of Computer Vision},
  volume={132},
  number={12},
  pages={5635--5662},
  year={2024},
  publisher={Springer}
}

@article{janet2019quantitative,
  title={A quantitative uncertainty metric controls error in neural network-driven chemical discovery},
  author={Janet, Jon Paul and Duan, Chenru and Yang, Tzuhsiung and Nandy, Aditya and Kulik, Heather J},
  journal={Chemical science},
  volume={10},
  number={34},
  pages={7913--7922},
  year={2019},
  publisher={Royal Society of Chemistry}
}

@article{musielewicz2024latent,
  title={Improved uncertainty estimation of graph neural network potentials using engineered latent space distances},
  author={Musielewicz, Joseph and Lan, Janice and Uyttendaele, Matt and Kitchin, John R},
  journal={The Journal of Physical Chemistry C},
  volume={128},
  number={49},
  pages={20799--20810},
  year={2024},
  publisher={ACS Publications}
}

@article{bigi2024prediction,
  title={A prediction rigidity formalism for low-cost uncertainties in trained neural networks},
  author={Bigi, Filippo and Chong, Sanggyu and Ceriotti, Michele and Grasselli, Federico},
  journal={Machine Learning: Science and Technology},
  volume={5},
  number={4},
  pages={045018},
  year={2024},
  publisher={IOP Publishing}
}

@article{kellner2024shallow,
  title={Uncertainty quantification by direct propagation of shallow ensembles},
  author={Kellner, Matthias and Ceriotti, Michele},
  journal={Machine Learning: Science and Technology},
  volume={5},
  number={3},
  pages={035006},
  year={2024},
  publisher={IOP Publishing}
}

@article{bilbrey2025foundationuq,
  title={Uncertainty quantification for neural network potential foundation models},
  author={Bilbrey, Jenna A and Firoz, Jesun S and Lee, Mal-Soon and Choudhury, Sutanay},
  journal={npj Computational Materials},
  volume={11},
  number={1},
  pages={109},
  year={2025},
  publisher={Nature Publishing Group UK London}
}

@article{perez2025misspecified,
  title={Uncertainty quantification for misspecified machine learned interatomic potentials},
  author={Perez, Danny and Subramanyam, Aparna PA and Maliyov, Ivan and Swinburne, Thomas D},
  journal={npj Computational Materials},
  volume={11},
  number={1},
  pages={263},
  year={2025},
  publisher={Nature Publishing Group UK London}
}

@article{grasselli2025uqera,
  title={Uncertainty in the era of machine learning for atomistic modeling},
  author={Grasselli, Federico and Chong, Sanggyu and Kapil, Venkat and Bonfanti, Silvia and Rossi, Kevin},
  journal={Digital Discovery},
  volume={4},
  number={10},
  pages={2654--2675},
  year={2025},
  publisher={Royal Society of Chemistry}
}

@article{alain2016probing,
  title={Understanding intermediate layers using linear classifier probes},
  author={Alain, Guillaume and Bengio, Yoshua},
  journal={arXiv preprint arXiv:1610.01644},
  year={2016}
}

@article{fannjiang2022conformal,
  title={Conformal prediction under feedback covariate shift for biomolecular design},
  author={Fannjiang, Clara and Bates, Stephen and Angelopoulos, Anastasios N and Listgarten, Jennifer and Jordan, Michael I},
  journal={Proceedings of the National Academy of Sciences},
  volume={119},
  number={43},
  pages={e2204569119},
  year={2022},
  publisher={National Academy of Sciences}
}

@article{kurniawan2025comparative,
  title={Comparative study of ensemble-based uncertainty quantification methods for neural network interatomic potentials},
  author={Kurniawan, Yonatan and Wen, Mingjian and Tadmor, Ellad B and Transtrum, Mark K},
  journal={arXiv preprint arXiv:2508.06456},
  year={2025}
}

@article{landrum2013rdkit,
  title={RDKit: A software suite for cheminformatics, computational chemistry, and predictive modeling},
  author={Landrum, Greg and others},
  journal={Greg Landrum},
  volume={8},
  number={31.10},
  pages={5281},
  year={2013}
}

@misc{2604.24607,
Author = {Matthias Kellner and Teitur Hansen and Thomas Bligaard and Karsten Wedel Jacobsen and Michele Ceriotti},
Title = {Errors that matter: Uncertainty-aware universal machine-learning potentials calibrated on experiments},
Year = {2026},
Eprint = {arXiv:2604.24607},
}

\end{document}

% ── Manual SI header ─────────────────────────────────────────────────────────
\begin{center}
    {\Large\textbf{Supplementary Information}}\\[0.8em]
    {\large\textbf{\papername}}\\[0.5em]
    Shams Mehdi, Ilkwon Cho, Olexandr Isayev$^*$\\[0.2em]
    {\small Department of Chemistry, Mellon College of Science,
    Carnegie Mellon University, Pittsburgh, Pennsylvania 15213, USA}\\[0.2em]
    {\small $^*$\texttt{olexandr@olexandrisayev.com}}
\end{center}

\vspace{1em}
\noindent\rule{\textwidth}{0.4pt}
\vspace{1em}

% ── Supplementary Figures ────────────────────────────────────────────────────
\clearpage
\section*{Supplementary Figures}

\begin{figure}[h]
    \centering
    \includegraphics[width=\textwidth]{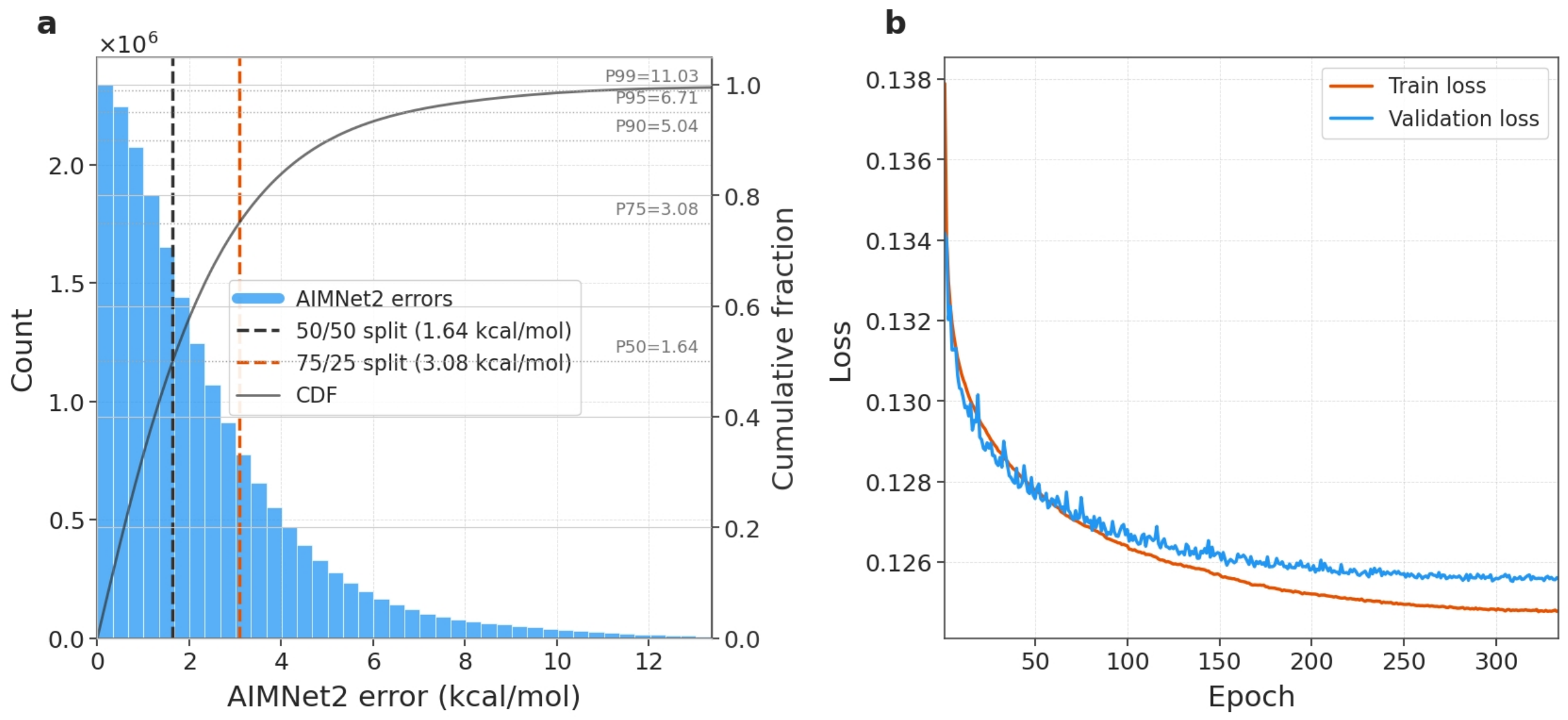}
    \caption{
        \textbf{AIMNet2 error distribution and training dynamics.}
        \textbf{(a)} Distribution of per-molecule AIMNet2 absolute errors (kcal/mol)
        with cumulative distribution function (right axis, black curve).
        Dashed vertical lines indicate the class boundary thresholds used for
        {\methodname} training; percentile markers are shown for reference.
        \textbf{(b)} Training (orange) and validation (blue) loss curves across
        epochs.
    }
    \label{fig:si_aimnet2_error}
\end{figure}

\begin{figure}[h]
    \centering
    \includegraphics[width=\textwidth]{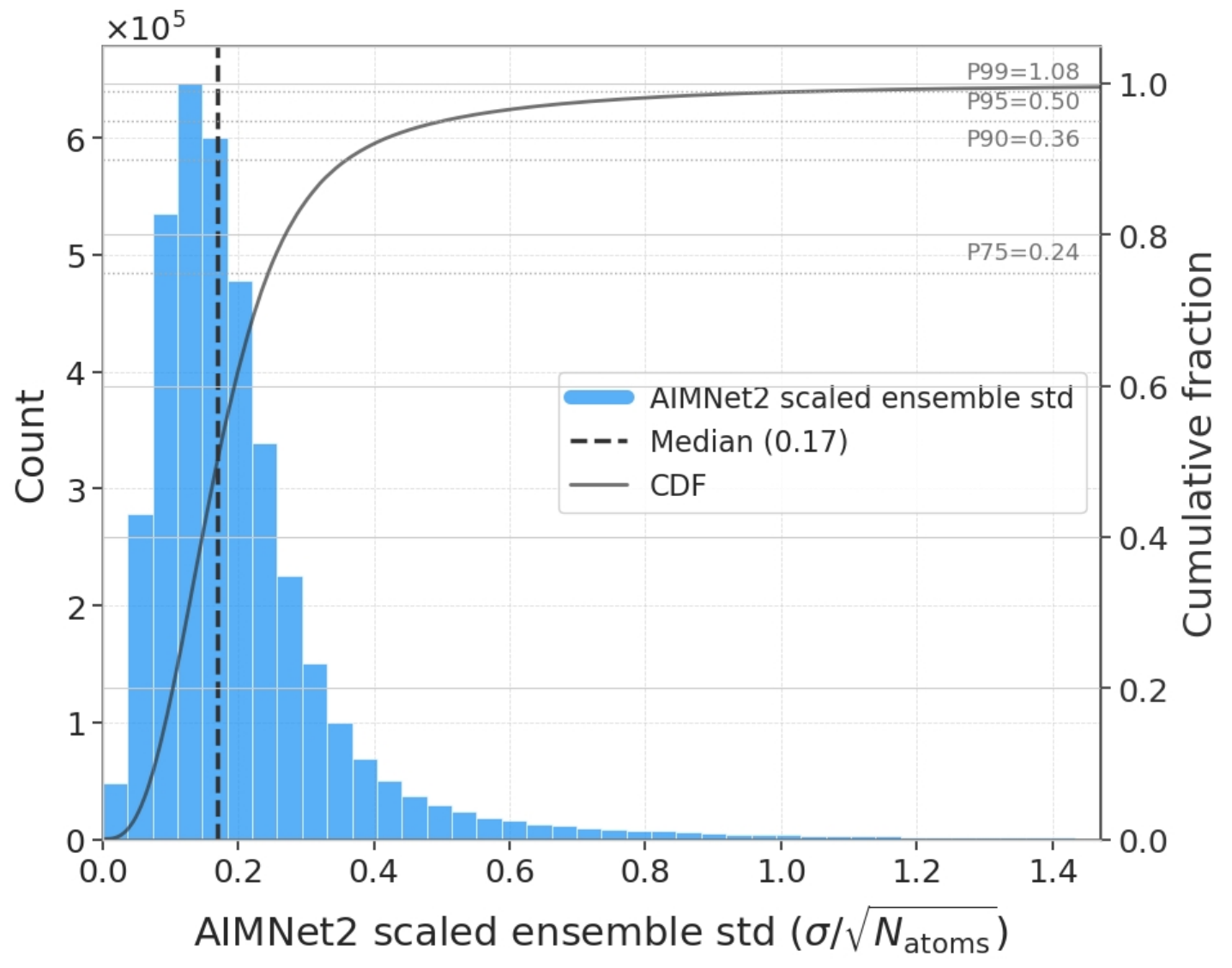}
    \caption{\textbf{Distribution of scaled ensemble standard deviation for AIMNet2.} Histogram of $\frac{\sigma} {\sqrt{N_\mathrm{atoms}}}$, where $\sigma$ (kcal/mol) is the standard deviation across a 4-model AIMNet2 ensemble and $N_\mathrm{atoms}$ is the number of atoms in a molecule, evaluated on the 3.76M held-out test set. The dashed vertical line marks the median. The right axis shows the cumulative distribution function (CDF) with percentile annotations.}

    \label{fig:si_aimnet2_ensemble_error}
\end{figure}

\begin{figure}[h]
    \centering
    \includegraphics[width=\textwidth]{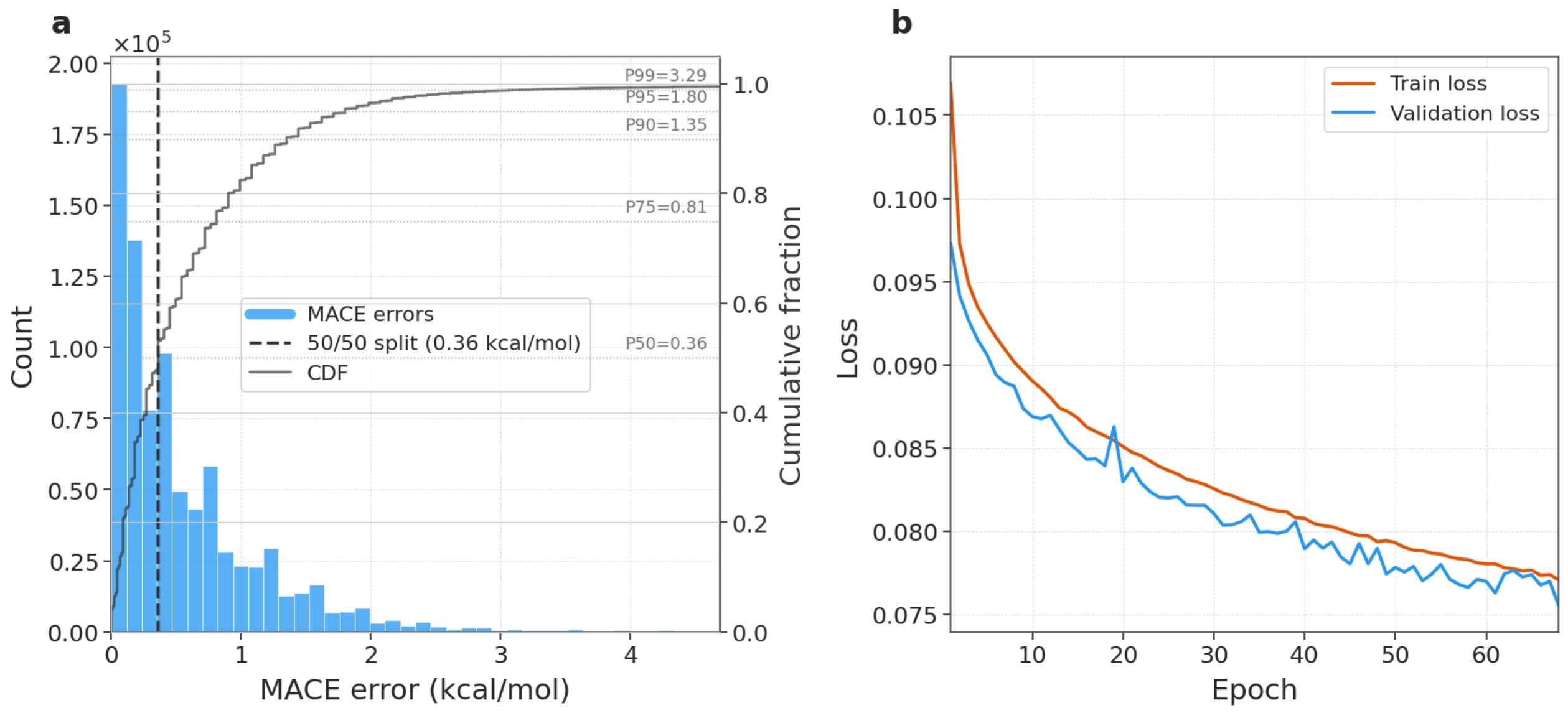}
    \caption{\textbf{MACE-OFF23 error distribution and {\methodname} training curves.}
    (a) Distribution of MACE-OFF23 absolute prediction errors on the 855,905-molecule training set. The dashed vertical line marks the 50/50 class boundary at 0.36~kcal/mol. The right axis shows the cumulative distribution function (CDF) with percentile annotations.
    (b) {\methodname} train and validation loss as a function of epoch. The validation loss closely tracks the training loss throughout, indicating no overfitting. Here, we implemented early stopping at epoch 68.}
    \label{fig:mace_si}
\end{figure}

\begin{figure}[h]
\centering
\includegraphics[width=\textwidth]{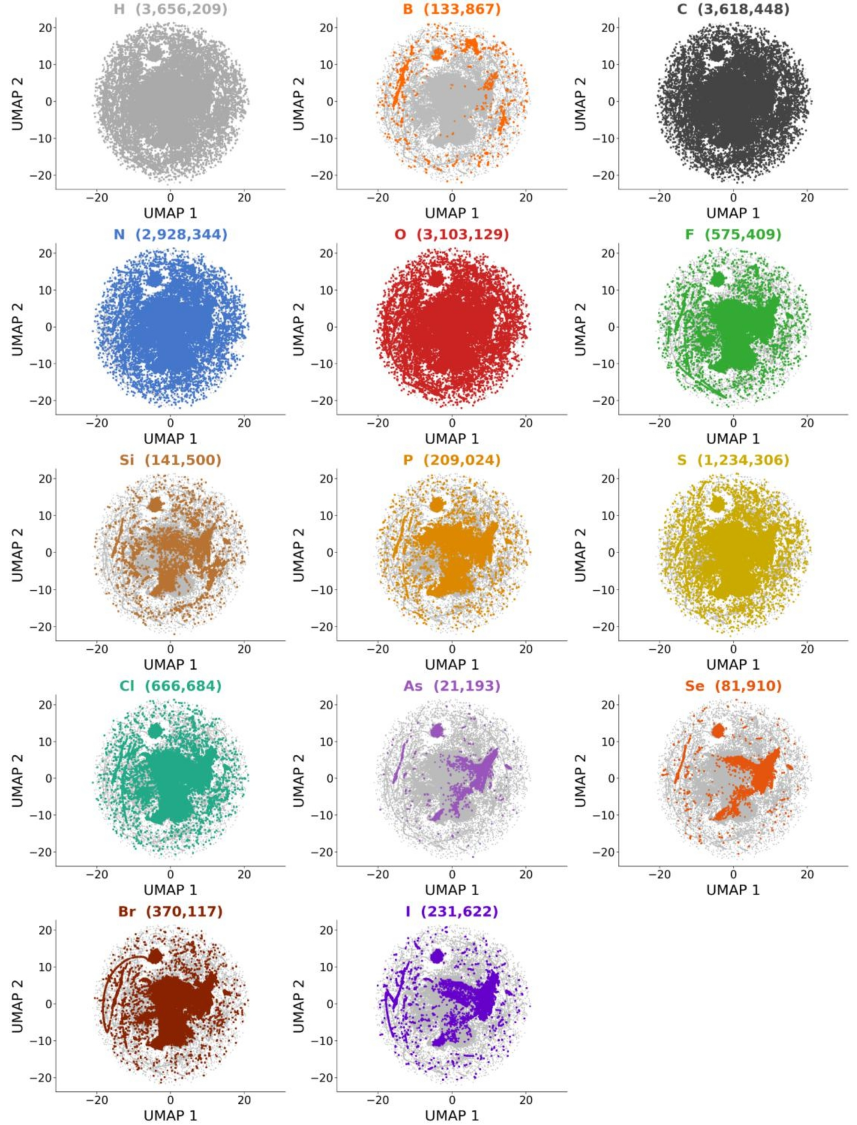}
\caption{\textbf{Element presence projected onto {\methodname} UMAP embedding for the 3.76M AIMNet2 test set.}
Each panel shows molecules containing at least one atom of the indicated element (colored)
overlaid on the full dataset (gray). Molecule counts are given in parentheses.
Rare and heavy elements (B, Si, As, Se, Br, I) concentrate in spatially distinct
regions of the embedding that correspond to high-$P(\text{unreliable})$ clusters, while abundant elements (H, C, N, O) span the full manifold.
}\label{fig:umap_elements_aimnet}
\end{figure}

\begin{figure}[h]
\centering
\includegraphics[width=\textwidth]{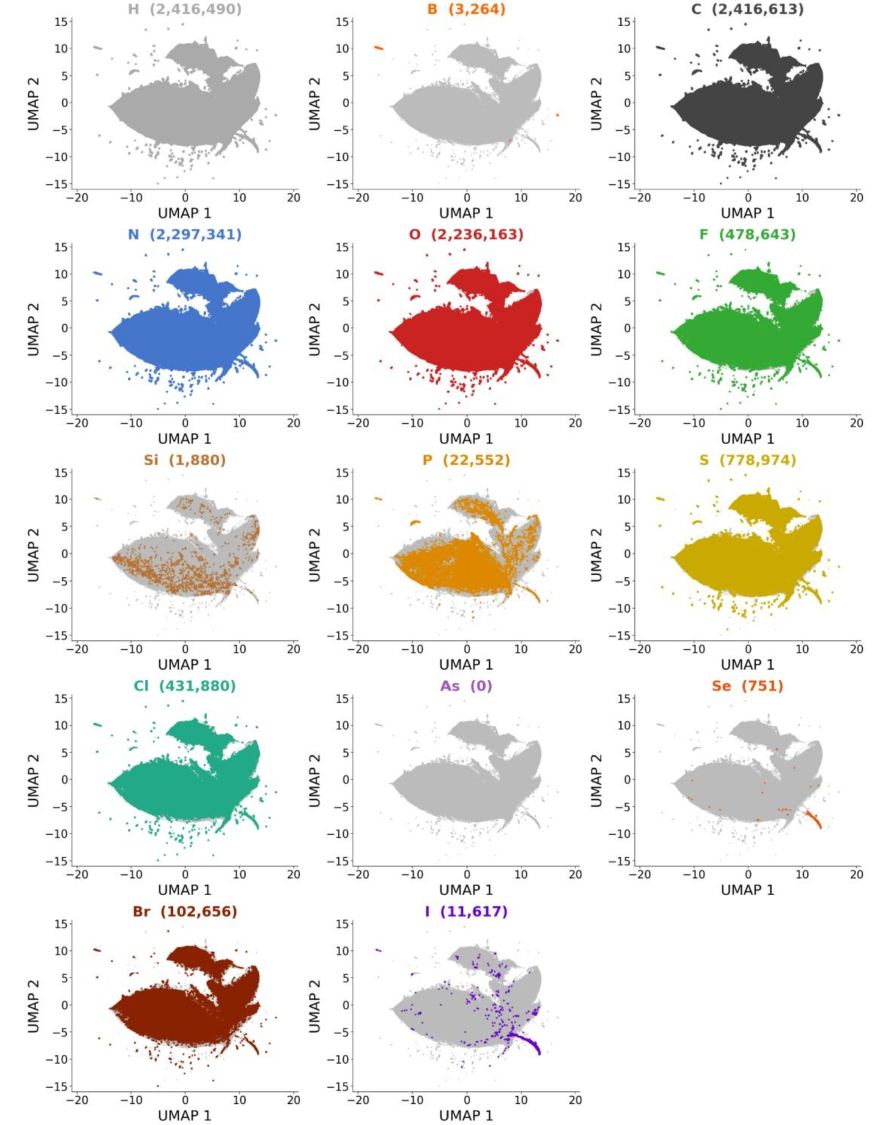}
\caption{\textbf{Element presence projected onto {\methodname} UMAP embedding for the 2.4M Reidenbach et al.\ dataset.}
Same layout as Fig.~S\ref{fig:umap_elements_aimnet}.
The dataset is dominated by CHNO chemistry.
Iodine (11,617) and selenium (751) map onto the unreliable tail while boron clusters at a distant island, consistent with the AIMNet2 test set analysis.
Bromine (102,656) is more broadly distributed, occupying both reliable and unreliable
regions.
}\label{fig:umap_elements_nikitin}
\end{figure}

\begin{figure}[h]
\centering
\includegraphics[width=\textwidth]{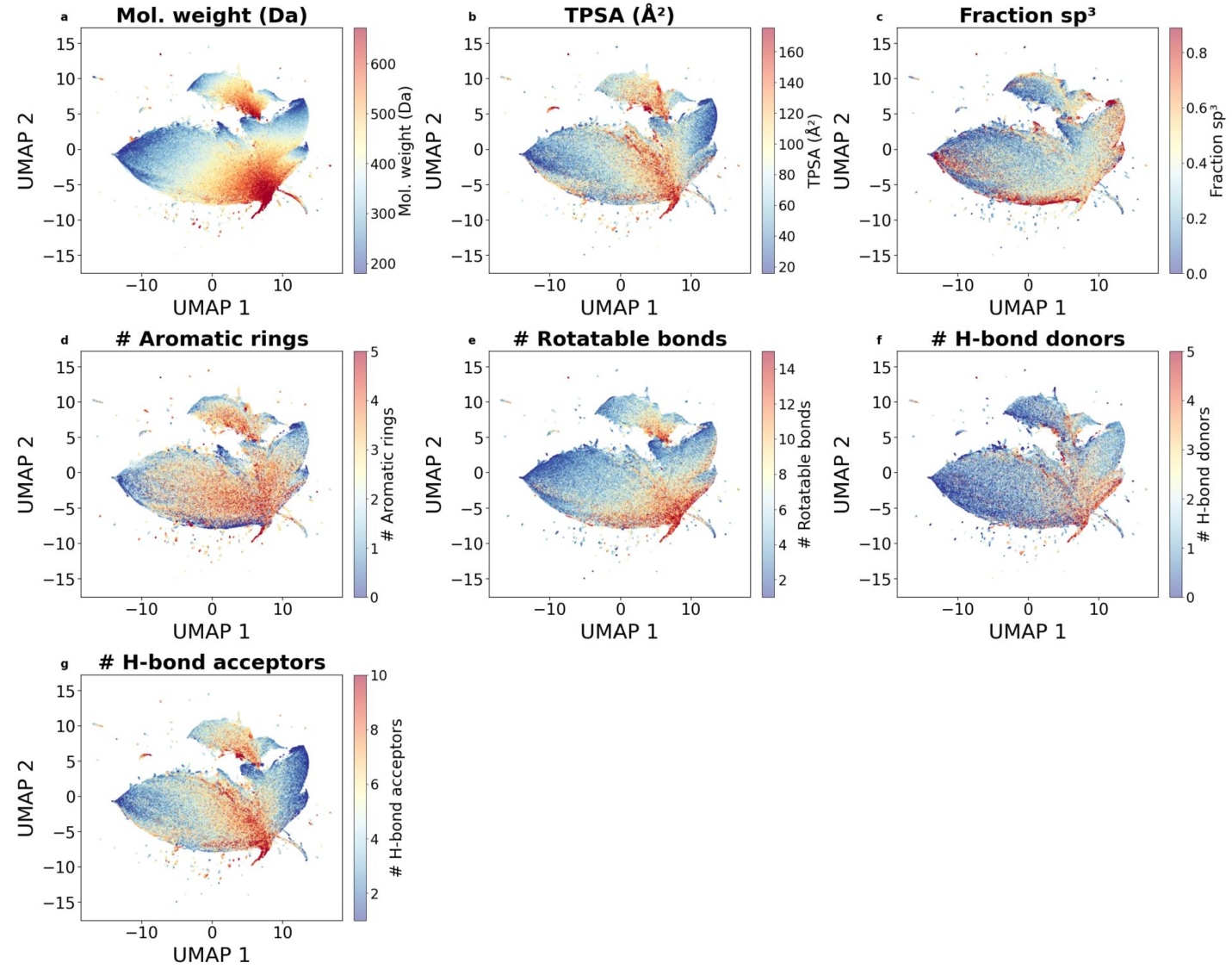}
\caption{\textbf{RDKit molecular descriptor projections onto {\methodname} UMAP embedding for the 2.4M Reidenbach et al.\ dataset.}
Descriptors were computed from connectivity information using RDKit.
Molecular weight, TPSA, number of H-bond acceptors, and number of rotatable bonds
all increase toward the unreliable tail of the embedding, indicating that large, polar, and flexible molecules
are more likely to receive unreliable AIMNet2 predictions.
Fraction sp$^3$ shows the different trend, highly saturated molecules are
concentrated in the reliable core.
Number of aromatic rings and H-bond donors show weaker spatial correlation
with the reliability score.
}\label{fig:umap_props_nikitin}
\end{figure}

\begin{figure}[h]
\centering
\includegraphics[width=\textwidth]{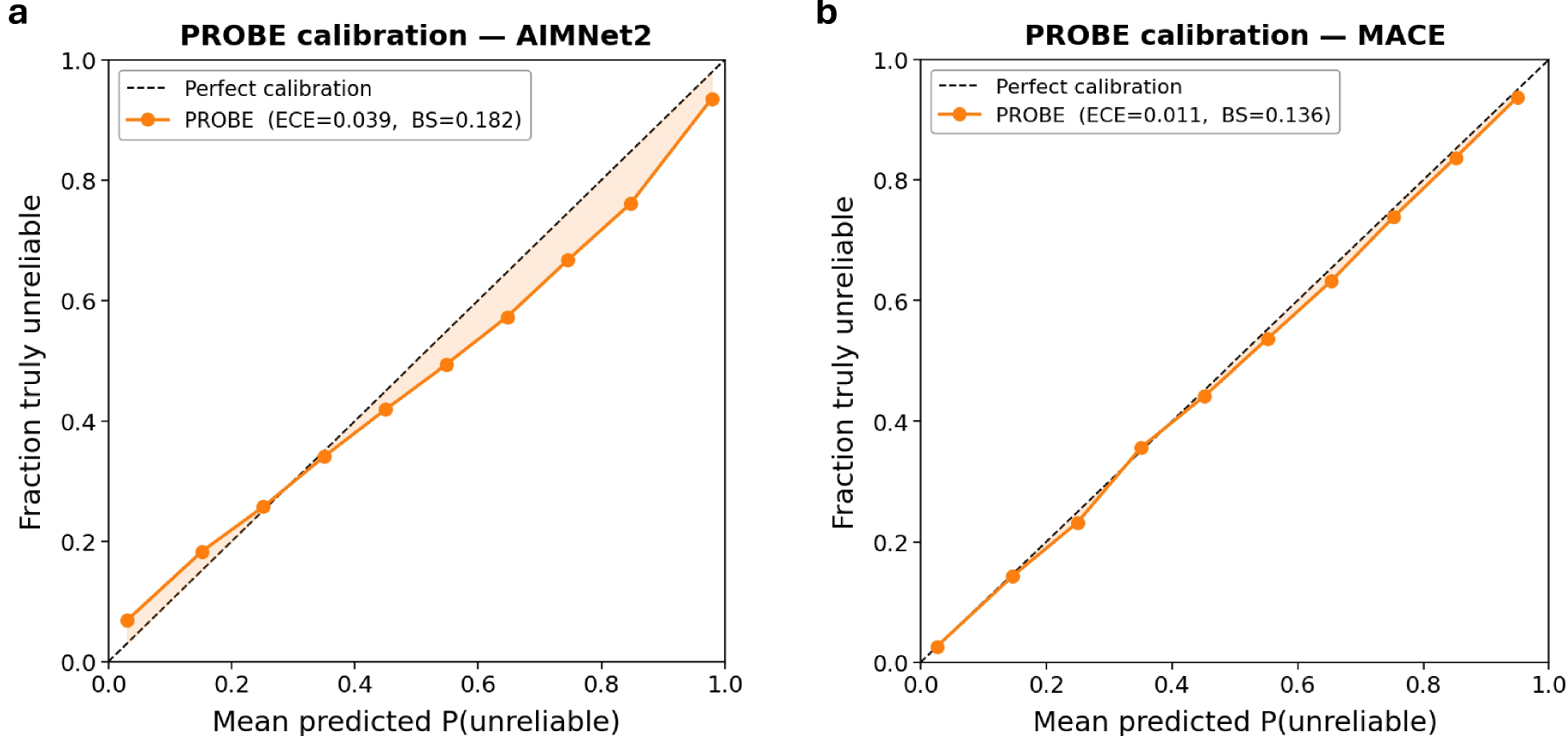}
\caption{\textbf{{\methodname} probability calibration.}
Reliability diagrams for {\methodname} on (a) AIMNet2 (3.76M held-out molecules)
and (b) MACE-OFF23 (50,195 held-out molecules). Each point shows the fraction of
molecules truly unreliable within a bin of predicted $P(\text{unreliable})$; the
dashed diagonal indicates perfect calibration. Shading shows the gap between the
calibration curve and the diagonal. ECE = expected calibration error; BS = Brier
score.
}\label{fig:calibration}
\end{figure}

% ── Supplementary Tables ─────────────────────────────────────────────────────

% \section*{Supplementary Tables}
% (add supplementary tables here as needed)